\documentclass[acmtog, screen]{acmart}

\usepackage{makecell}
\usepackage{enumitem}
\usepackage{xspace}
\usepackage{colortbl}
\usepackage{amsfonts}
\usepackage{pifont}
\usepackage{nicefrac}
\usepackage{multirow}
\usepackage{textcomp}

\definecolor{lightblue}{RGB}{102, 178, 255}
\newcommand{\revise}[1]{\textcolor{lightblue}{#1}}
\newcommand{\note}[1]{\textcolor{blue}{#1}}

\renewcommand{\revise}[1]{#1}
\renewcommand{\note}[1]{}

\copyrightyear{2025} 
\acmYear{2025}
\setcopyright{acmlicensed}\acmConference[SA Conference Papers '25]{SIGGRAPH Asia
2025 Conference Papers}{December 15--18, 2025}{Hong Kong, Hong Kong}
\acmBooktitle{SIGGRAPH Asia 2025 Conference Papers (SA Conference Papers '25),
December 15--18, 2025, Hong Kong, Hong Kong}
\acmDOI{10.1145/3757377.3763878}
\acmISBN{979-8-4007-2137-3/2025/12}

\begin{document}
\title{EGG-Fusion: Efficient 3D Reconstruction with Geometry-aware Gaussian Surfel on the Fly}

\author{Xiaokun Pan}
\email{panxkun@gmail.com}
\orcid{0000-0002-7438-1665}
\affiliation{%
 \institution{State Key Lab of CAD\&CG, Zhejiang University}
 \city{Hangzhou}
 \country{China}
}

\author{Zhenzhe Li}
\email{12321093@zju.edu.cn}
\orcid{0009-0000-4008-4665}
\affiliation{%
 \institution{State Key Lab of CAD\&CG, Zhejiang University}
 \city{Hangzhou}
 \country{China}
}

\author{Zhichao Ye}
\email{11721010@zju.edu.cn}
\orcid{0000-0001-7304-5646}
\affiliation{%
 \institution{SenseTime Research}
 \city{Hangzhou}
 \country{China}
}
\authornotemark[1]

\author{Hongjia Zhai}
\email{zhj1999@zju.edu.cn}
\orcid{0000-0002-7729-8787}
\affiliation{%
 \institution{State Key Lab of CAD\&CG, Zhejiang University}
 \city{Hangzhou}
 \country{China}
}

\author{Guofeng Zhang}
\email{zhangguofeng@zju.edu.cn}
\orcid{0000-0001-5661-8430}
\affiliation{%
 \institution{State Key Lab of CAD\&CG, Zhejiang University}
 \city{Hangzhou}
 \country{China}
}
\authornote{Guofeng Zhang and Zhichao Ye are Corresponding Authors}

\begin{abstract}
\label{sec:abstract}
Real-time 3D reconstruction is a fundamental task in computer graphics.
Recently, differentiable-rendering-based SLAM system has demonstrated significant potential, enabling photorealistic scene rendering through learnable scene representations such as Neural Radiance Fields (NeRF) and 3D Gaussian Splatting (3DGS). 
Current differentiable rendering methods face dual challenges in real-time computation and sensor noise sensitivity, leading to degraded geometric fidelity in scene reconstruction and limited practicality.
To address these challenges, we propose a novel real-time system EGG-Fusion, featuring robust sparse-to-dense camera tracking and a geometry-aware Gaussian surfel mapping module, introducing an information filter-based fusion method that explicitly accounts for sensor noise to achieve high-precision surface reconstruction. 
The proposed differentiable Gaussian surfel mapping effectively models multi-view consistent surfaces while enabling efficient parameter optimization.
Extensive experimental results demonstrate that the proposed system achieves a surface reconstruction error of 0.6\textit{cm} on standardized benchmark datasets including Replica and ScanNet++, representing over 20\% improvement in accuracy compared to state-of-the-art (SOTA) GS-based methods. 
Notably, the system maintains real-time processing capabilities at 24 FPS, establishing it as one of the most accurate differentiable-rendering-based real-time reconstruction systems. 
Project Page: \url{https://zju3dv.github.io/eggfusion/}.
\end{abstract}

%
%

\ccsdesc[300]{Computing methodologies~Point-based models}
\ccsdesc[300]{Computing methodologies~Rasterization}
\ccsdesc[500]{Computing methodologies~Reconstruction}
\ccsdesc[500]{Computing methodologies~Tracking}

\keywords{3D Reconstruction, SLAM, 3D Gaussian Splatting, Gaussian Surfels, Differential Rendering}
  
\begin{teaserfigure}
\includegraphics[width=\textwidth, trim={0cm 2.2cm 0cm 0.cm}]{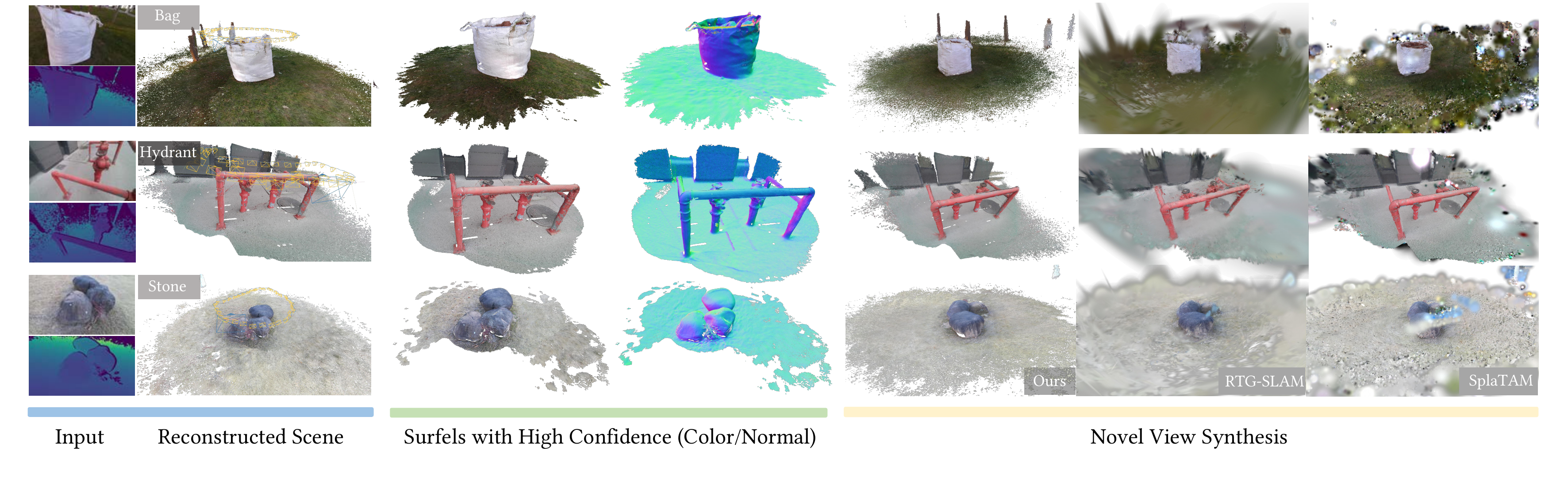}
\caption{
Qualitative comparison on three real-world scenes reconstructed by our system and the state-of-the-art Gaussian Splatting based SLAM methods (RTG-SLAM\cite{rtg_slam}, SplaTAM\cite{splatam}). 
Leveraging Gaussian surface representation and surface fusion with information filter, our approach achieves higher geometric accuracy and rendering fidelity (yellow bar). 
Moreover, our method supports the extraction of high-confidence scene surfaces by maintaining the information matrix of scene primitives (green bar).
} 
\label{fig:teaser}
\end{teaserfigure}

\maketitle
\section{Introduction}
\label{sec:introduction}

Real-time 3D reconstruction technology aims to generate high-precision 3D models by efficiently processing real-world data, 
which has been extensively studied in the fields of computer graphics and has demonstrated broad application prospects in fields such as mixed reality, autonomous driving, and robotics.

Over the past decades, significant progress has been made in RGBD-based Simultaneous Localization and Mapping (SLAM) ~\cite{newcombeDTAMDenseTracking2011a,newcombeKinectFusionRealtimeDense2011, whelanElasticFusionDenseSLAM2015c, daiBundleFusionRealtimeGlobally2017c, xuHRBFFusionAccurate3D2022c,kerlDenseVisualSLAM2013,zhouDenseSceneReconstruction2013,newcombeDynamicFusionReconstructionTracking2015}. 
Meanwhile, researchers have proposed various scene representations to better model scenes and improve reconstruction efficiency, such as point clouds~\cite{kellerRealTime3DReconstruction2013,zhouDenseSceneReconstruction2013, kerlDenseVisualSLAM2013}, surfels~\cite{whelanElasticFusionDenseSLAM2015c, xuHRBFFusionAccurate3D2022c}, and Truncated Signed Distance Fields (TSDF)~\cite{newcombeKinectFusionRealtimeDense2011, newcombeDynamicFusionReconstructionTracking2015, daiBundleFusionRealtimeGlobally2017c}. 
However, these map representations have limited expressive capacity and struggle to handle complex scenes, resulting in limitations in representing precise geometric details and scalability.

In recent years, the rapid development of differentiable rendering techniques has significantly enhanced the realism and visual quality of 3D reconstruction, with notable advancements in technologies such as  NeRF~\cite{mildenhallNeRFRepresentingScenes2020}  and 3DGS~\cite{kerbl3DGaussianSplatting2023}.
Existing studies~\cite{lu2024scaffold, yu2024mip, zhai2025neuraloc, zhai2024nis} have demonstrated that differentiable rendering techniques can be effectively integrated into SLAM frameworks~\cite{qinVINSMonoRobustVersatile2018c,liRDVIORobustVisualInertial2024, pan2024robust, pan2025msslam, chen2025dwvio}.
In particular, real-time reconstruction methods based on 3DGS~\cite{splatam, rtg_slam} have shown great promise due to their efficient differentiable rasterization capabilities.
However, the high degree of freedom in the parameterization of 3DGS primitive introduces geometric ambiguities, which can degrade reconstruction accuracy.
This issue is further exacerbated in some complex scenes by factors such as limited viewpoint coverage and sensor noise ~\cite{cao2018real, fontan2020information,fontan2023sid}, leading to significant reconstruction errors or even complete failure.
These challenges pose serious limitations to the reliability and scalability of 3DGS-based real-time reconstruction methods in practical applications. 

To address these issues, we propose EGG-Fusion, a novel framework for high-quality and real-time 3D reconstruction in real-world environments.
Our approach begins by adopting Gaussian surfels~\cite{huang2DGaussianSplatting2024, daiHighqualitySurfaceReconstruction2024} as the scene representation, enabling consistent multi-view geometric modeling. 
However, Gaussian surfels alone are insufficient to improve reconstruction quality, as sensor noise remains a pervasive challenge.
To address this, we introduce a novel surfel fusion strategy with information filter~\cite{thrun2004simultaneous, maybeck1982stochastic}, which incrementally updates the geometric attributes of surfels using depth observations from each frame. 
This incremental fusion process jointly refines the surfel geometry and its associated information matrix, effectively enhancing the stability and fidelity of the reconstruction. The information matrix also facilitates confidence estimation for each primitive, allowing the system to extract highly reliable reconstruction results.
Thanks to frame-wise geometric updates, the scene maintains high-precision geometry throughout reconstruction. 
Consequently, during map optimization, only minor refinements are required, leading to rapid convergence with significantly reduced computational overhead. 
To enable robust camera pose estimation, we design a sparse-to-dense tracking strategy that combines the robustness of sparse features with the precision of dense alignment.
Additionally, we propose a geometry-aware surfel initialization method, which adaptively determines the density of scene primitives based on depth observations. This strategy ensures accurate surface alignment while producing a compact and efficient scene representation.
Experimental results demonstrate that EGG-Fusion operates in real time at 24 FPS, and consistently outperforms SOTA methods in terms of tracking accuracy, reconstruction quality, and rendering fidelity. As shown in Fig.~\ref{fig:teaser}, our method maintains high geometric precision and visual quality, even when tested under challenging, unseen viewpoints.

In summary, the main contributions of our work are as follows:
\begin{itemize}[leftmargin=2em]
\item We propose EGG-Fusion, a Gaussian surfel-based real-time 3D reconstruction system for high-quality scene modeling in real-world environments. We release the source code to facilitate research reproducibility and community advancement.
\item We propose an information-filter-based surfel fusion method that achieves stable high-precision reconstruction by dynamically updating geometric information matrices, while significantly accelerating optimization convergence.
\item Extensive experimental results demonstrate that the proposed method outperforms current GS-based SOTA methods in tracking accuracy, reconstruction quality, and rendering quality, while running in real-time at 24 FPS. 
\end{itemize}
\begin{figure*}[t!]
  \centering
  \includegraphics[clip,width=0.95\linewidth, trim={0.0cm 18.2cm 19cm 0cm}]{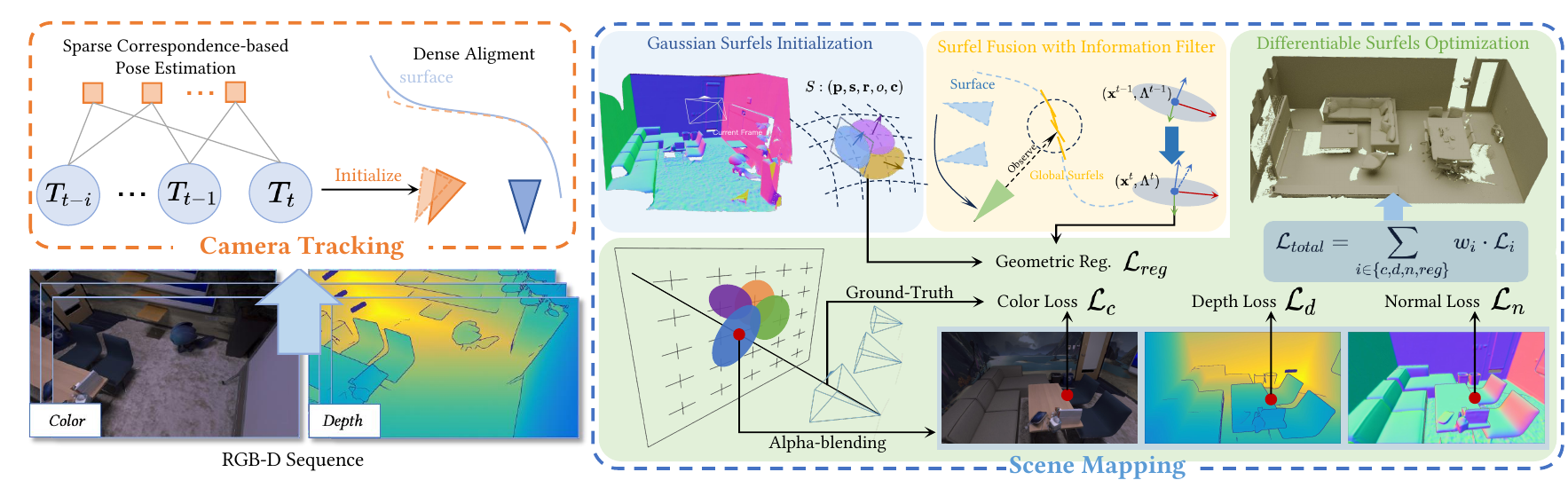}
  \caption{
Framework of EGG-Fusion.
Our framework is divided into two integral components. 
In the scene mapping module (Sec.~\ref{sec:scene_mapping}), Gaussian surfels are utilized as the fundamental primitives for scene representation and can achieve high-quality real-time reconstruction
The camera tracking module (Sec.~\ref{sec:pose_estim}) employs a sparse-to-dense strategy to ensure robust estimation of camera poses.
}
  \label{fig:framework}
\end{figure*}

\section{Related Work}
\label{sec:relatedwork}


\subsection{Dense Visual SLAM}
In classical dense visual SLAM frameworks, DTAM~\cite{newcombeDTAMDenseTracking2011a} pioneered the dense 3D model reconstruction of an indoor scene from monocular video by directly tracking the camera to the model using photometric error, making it robust to rapid camera movement and defocus blur.~\cite{newcombeKinectFusionRealtimeDense2011, newcombeDynamicFusionReconstructionTracking2015} utilizes consumer-grade RGB-D cameras (Microsoft Kinect) to accomplish featureless camera tracking by optimizing the transformation of depth information to the TSDF target.
~\cite{kellerRealTime3DReconstruction2013} proposed a global map representation scheme based on point with radius (surfels), enabling reconstruction in dynamic scenes. 
BundleFusion~\cite{daiBundleFusionRealtimeGlobally2017c} achieves robust pose estimation through an efficient hierarchical approach and attains high-quality real-time reconstruction results using a parallel optimization framework.
With the assistance of neural networks, hand-crafted components in classic frameworks are replaced by end-to-end network architectures. ~\cite{koestlerTANDEMTrackingDense2021b, teedDROIDSLAMDeepVisual2021, minVOLDORVisualOdometry2020} employ prior-based dense depth or optical flow to perform direct image alignment, making photometric/geometric bundle adjustment (BA) within a keyframe sliding window feasible.
Compared to traditional approaches, end-to-end architectures~\cite{matsukiCodeMappingRealTimeDense2021,bloeschCodeSLAMLearningCompact2019,zhiSceneCodeMonocularDense2019} simplify the problem's modeling complexity but demand significantly higher computational resources. In addition, these approaches are still constrained by traditional scene representations, lacking flexibility and scalability.

\subsection{Differentiable-based Scene Representation}
Since the introduction of NeRF~\cite{mildenhallNeRFRepresentingScenes2020}, neural-based differentiable rendering has emerged as a promising approach for scene representation.
A series of studies have explored improvements in rendering quality~\cite{barron2021mip, barron2022mip}, joint pose optimization~\cite{linBARFBundleAdjustingNeural2021, wangNeRFNeuralRadiance2022}, rendering speed~\cite{mullerInstantNeuralGraphics2022, sunDirectVoxelGrid2022, fridovich-keilPlenoxelsRadianceFields2022a}, and reconstruction quality~\cite{wangNeuSLearningNeural2023, li2023neuralangelo}. However, challenges remain in terms of slow optimization convergence for scene appearance/geometry and bottlenecks in rendering speed. 
3DGS represents a scene using a set of ellipsoids, significantly enhancing the rendering performance of differentiable scene representations by rasterizing these 3D ellipsoids directly in 2D image space. 
Subsequent works have extended its capabilities, achieving advancements in rendering quality~\cite{yu2024mip, lu2024scaffold}, geometry reconstruction~\cite{guedon2024sugar}, and performance on large-scale scenes~\cite{liu2025citygaussian, kerbl2024hierarchical}. The variant 2D Gaussian surfels~\cite{daiHighqualitySurfaceReconstruction2024, huang2DGaussianSplatting2024} constrains the z-axis of 3D ellipsoids to enable precise geometric reconstruction of the scene.

\subsection{Differentiable Rendering based SLAM}
The differentiable property in system-level was first discussed in $\bigtriangledown $SLAM~\cite{jatavallabhulaGradSLAMAutomagicallyDifferentiable2020}, which framed the entire system as a differentiable computation graph. 
However, it retained traditional 3D representations. 
iMap~\cite{sucarIMAPImplicitMapping2021} was the first to employ differentiable scene representations for SLAM system, leading to a series of subsequent works, including those utilizing hierarchical grids~\cite{nice_slam}, voxels~\cite{vox_fusion}, hybrid implicit parameters and grids~\cite{co_slam}, and point-based features~\cite{point_slam}. These methods leverage differentiable scene representations for online mapping but suffer from significant efficiency issues, impacting their real-time performance and scalability.
More recently, SLAM systems~\cite{mono_gs,splatam,yanGSSLAMDenseVisual2024,rtg_slam} based on 3DGS have demonstrated impressive performance in novel view synthesis and rendering speed, thanks to the efficiency of the scene representation. However, this discrete primitive-based representation has potential limitations in terms of parameter efficiency. 
\section{Methodology}
\label{sec:method}

As illustrated in Fig.~\ref{fig:framework}, the system is composed of two modules: camera tracking and scene mapping. 
This section first introduces the definition of Gaussian surfels and the associated notation (Sec.~\ref{sec:preliminary}). 
Subsequently, the proposed scene mapping module is elaborated, and the approach to achieving high-quality scene surface reconstruction using Gaussian surfels as the scene representation is discussed (Sec.~\ref{sec:scene_mapping}). 
Finally, a sparse-to-dense camera tracking strategy is proposed (Sec.~\ref{sec:pose_estim}) to achieve robust and efficient real-time pose estimation.

\subsection{Preliminary}
\label{sec:preliminary}

\subsubsection{2D Gaussian Surfels.}
The Gaussian surfel is a primitive for learnable scene representation, with a spatial distribution approximated by a disk.
Each 2D Gaussian surfel can be represented by the attributes consisting of its center $\textbf{p}_i \in \mathbb{R}^3$, the scales of its two orthogonal Gaussian ellipse axes $\textbf{s}_i \in \mathbb{R}^2$, its rotation in quaternion form $\textbf{r}_i \in \mathbb{R}^4$ (in the global coordinate system), its opacity $o_i \in \mathbb{R}$, and its color $\textbf{c}_i \in \mathbb{R}^k$.
The color attribute is explicitly encoded as coefficients of spherical harmonic basis functions, where the dimension $k$ depends on the defined order, enabling the modeling of non-Lambertian surfaces. 
In our system, the scene can be represented as a collection of Gaussian surfels, which can be parameterized as $\mathcal{S} = \left \{ S_i: (\textbf{p}_i, \textbf{s}_i, \textbf{r}_i, o_i, \textbf{c}_i) \right\}_{i=0}^{n}.$

\subsubsection{Differentiable Gaussian Splatting.}
In the differentiable rendering pipeline, each Gaussian surfel $S_i$ is transformed into image space based on the camera pose, and its 2D covariance $\Sigma_i^{2D}$ in image space are computed. After sorting by depth, alpha compositing is used for blending to obtain the final color \(\hat{C}\) with the alpha blending weight $\alpha_i = S^{'}_i(u;u_i, \Sigma_{i}^{2D})o_i$:
\begin{align} 
  T_i = \prod_{j=0}^{i-1}(1 - \alpha_j), \ \ \ \hat{C} = \sum_{i=0}^{n}T_i \cdot \alpha_i \cdot \textbf{c}_i.
    \label{eq:alpha_blending}
  \end{align}
Similarly, the depth map \(\hat{D}\) and normal map \(\hat{N}\) are rendered by:
\begin{align} 
\hat{A} = \frac{1}{1- T_{n+1}}\sum_{i=0}^{n}T_i \cdot \alpha_i \cdot a_i(u),
\label{eq:geo_alpha_blending}
\end{align}
where $\hat{A}$ is the rendered 2D information (\textit{e.g.}, depth $D$ or normal $N$), and $a$ is the 3D property.

\subsection{Scene Mapping with Gaussian Surfels}
\label{sec:scene_mapping}

Given the current frame \(I_t = \{C_t, D_t\} \), which includes a RGB image \(C_t \in \mathbb{R}^{H\times W \times 3}\) and its corresponding depth map \(D_t \in \mathbb{R}^{H\times W}\), we pre-process the data using the intrinsic parameters of the camera \(\{f_x, f_y, c_x, c_y\}\) to obtain the normal map \(N_t \in \mathbb{R}^{H\times W \times 3}\) and the vertex map \(V_t \in \mathbb{R}^{H\times W \times 3}\) in the camera coordinate system:

\begin{align} 
V_t(\mathbf{u}) = D_t(\mathbf{u}) \cdot \mathbf{K}^{-1} \cdot \left [ \mathbf{u},1 \right ] ^{\top }, \ \ \ N_t = \frac{ \nabla_x V_t \times \nabla_y V_t}{\left | \nabla_x V_t \times \nabla_y V_t \right | } ,
\end{align}
where $\mathbf{K} \in \mathbb{R}^{3 \times 3}$ is the intrinsic matrix and $\textbf{u} \in \mathbb{R}^{2}$ is the pixel coordinate.
Through the camera tracking strategy (Sec.~\ref{sec:pose_estim} ), we acquire the camera-to-world pose \(\mathbf{T}_{i} \in \mathbb{R}^{4 \times 4}\), enabling us to transform the vertex map and the normal map into the global coordinate system as \(V_t^w\) and \(N_t^w\).

\subsubsection{Geometry-aware Surfel Initialization}

\label{sec:surfel_init}
The proposed system incrementally expands the map by adaptively integrating Gaussian surfels from incoming RGB-D frames. Departing from previous works~\cite{mono_gs, splatam} that uniformly sample new primitives in image space, we introduce a rendering-aware strategy that selectively activates surfels only in geometrically salient regions: (1) low-opacity zones, revealing reconstruction deficiencies, and (2) areas with positive depth disparity, indicating newly observed foreground geometry. 
This targeted placement ensures high-fidelity reconstruction while inherently suppressing the redundancy of uniform sampling paradigms.
Each newly added Gaussian surfel is initialized by sampling geometric properties from the input depth map, including the surfel center position \( \mathbf{p} = V_t^w(\mathbf{u}) \) and surfel normal \( \mathbf{n} = N_t^w(\mathbf{u}) \).



Regarding surfel scale initialization \(\textbf{s}\), previous methods~\cite{mono_gs, rtg_slam} employ a fixed value in 3D space. 
 For some Gaussian ellipsoids located far from the camera, their projected area in image space becomes significantly small, necessitating dense sampling to adequately cover the corresponding image patch.
 To address this, we propose an adaptive scale initialization strategy: surfels located at the camera’s far plane are assigned larger scales , determined by the following principle:
\begin{align} 
\mathbf{s} = [\alpha_s \cdot d / f_x, \alpha_s \cdot d / f_y], 
\label{eq:adaptive_scale_init}
\end{align}
where \( \alpha_s \) is a pixel scaling factor used to regulate the absolute scale across different sampling rates. 
The intuitive outcome is that surfels farther from the camera have larger scales, while their projections in the image space maintain a consistent distribution. 
\note{(Author: refer to Mip-Splatting)}
\revise{This is similar to the analysis of sampling rate in ~\cite{yu2024mip}.}
Experiments demonstrate that this compact surfel initialization can achieve higher rendering quality.

\subsubsection{Surfel Fusion with Information Filter}
\label{sec:surfel_fusion}
\begin{figure}
  \centering
  \includegraphics[clip,width=1.0\linewidth, trim={-0.5cm 12cm 18cm 0cm}]{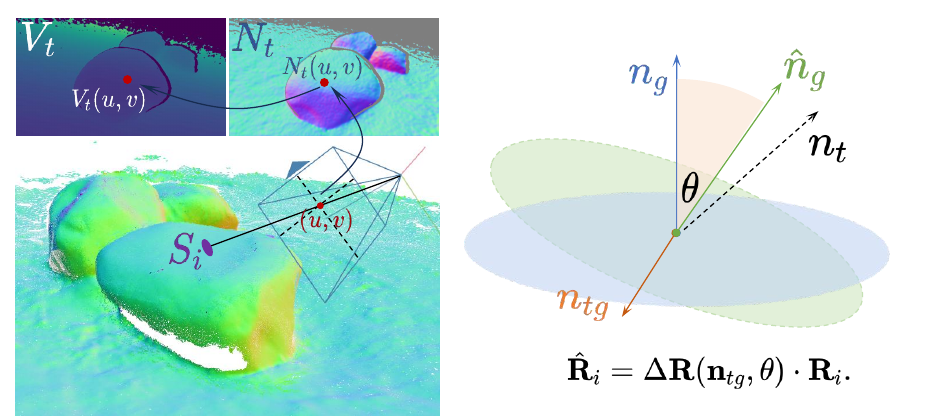}
  \caption{
  Gaussian surfels fusion.
  We ensure that surfels can be explicitly and continuously updated with new observations, enabling them to adhere to the scene surface (left), while new observations are utilized to update normal information (right), thereby achieving more accurate surface reconstruction. 
  }
  \label{fig:surfel_fusion}
\end{figure}

To mitigate the impact of depth noise from consumer-grade sensors, we propose an information-filter-based surface fusion method. 
The surfel state estimation is formulated as a Markov process, where the current state $\mathbf{x}^{t}$ depends solely on its previous state $\mathbf{x}^{t-1}$ and the latest sensor observation $I_t$.
For each input RGB-D frame, we perform recursive Bayesian updates on reobserved surfels to refine their geometric properties.  
Each surface element's geometric state is defined as $\mathbf{x}^{t}=[\mathbf{p},\mathbf{n}]^{\top}\in\mathbb{R}^{6}$ and is associated with a covariance matrix $\mathbf{{\Sigma}}^{t} \in \mathbb{R}^{6 \times 6}$ to quantify measurement uncertainty, enabling progressive confidence accumulation through sequential observations.
This probabilistic framework ensures continuous improvement in reconstruction quality, ultimately yielding high-fidelity surfaces.

Specifically, we can obtain the vertex and normal measurements $\mathbf{z}^t = [V_t(\mathbf{u}), N_t(\mathbf{u})]^{\top} \in \mathbb{R}^{6}$ based on the image location $\mathbf{u}$ of the surfel.
We construct the relationship between observation and state based on the state observation equation:
\begin{align}
\begin{aligned}
\mathbf{z}^t &= \mathbf{H} \mathbf{x}^t +\bar{\mathbf{t}} + \boldsymbol{\epsilon}, \quad \boldsymbol{\epsilon} \sim \mathcal{N}(0, \mathbf{{\Sigma}}_{\mathbf{z}}^t) ,
\\
\mathbf{H} 
&= 
\begin{bmatrix}
\mathbf{R} & \mathbf{0} \\
\mathbf{0} & \mathbf{R}
\end{bmatrix} \in \mathbb{R}^{6\times 6}, \quad
\bar{\mathbf{t}} = 
\begin{bmatrix}
\mathbf{t} \\
\mathbf{0} 
\end{bmatrix} \in \mathbb{R}^{6}.
\end{aligned} 
\end{align}
Here \textbf{H} denotes the observation matrix, which transforms the state variables from the global coordinate system into the camera coordinate system with current camera transformation $[\textbf{R},\textbf{t}]$.
$\boldsymbol{\epsilon}$ is the measurement noise, and $\mathbf{{\Sigma}}_{\mathbf{z}}^t$ corresponds to the associated covariance.
To ensure real-time computational efficiency, $\mathbf{{\Sigma}}_{\mathbf{z}}^t$ is simplified to a diagonal matrix represented in vector form as
$
\text{diag}(\mathbf{{\Sigma}}_{\mathbf{z}}^t) = [\sigma_p, \sigma_p, \sigma_p, \sigma_n, \sigma_n, \sigma_n]$ that $\sigma_p$ and $\sigma_n$ correspond to the position and normal respectively.
In practice, owing to the inherent characteristics of the sensor, the noise intensity correlates with depth, and thus 
$\sigma_p$ and $\sigma_n$ are proportional to the square of the depth.

We update the information matrix $\boldsymbol{\Lambda}$ and corresponding information vector matrix $\boldsymbol{\eta}$ using the following equations:
\begin{align}
\begin{matrix}
\boldsymbol{\Lambda}^t = \boldsymbol{\Lambda}^{t-1} + \mathbf{H}^\top \boldsymbol{\Lambda}^t_{\mathbf{z}} \mathbf{H},
\quad
\boldsymbol{\eta}^t = \boldsymbol{\eta}^{t-1} + \boldsymbol{\eta}^t_{\mathbf{z}},
 \\
\boldsymbol{\Lambda}^{t-1} = (\mathbf{{\Sigma}}^{t-1})^{-1},\ \ \boldsymbol{\eta}^{t-1} = \boldsymbol{\Lambda}^{t-1} \cdot \mathbf{x}^{t-1},
 \\
\mathbf{\Lambda}^t_{\mathbf{z}} = (\mathbf{{\Sigma}}^{t}_{\textbf{z}})^{-1},
\quad
\boldsymbol{\eta}^t_{\textbf{z}} = \mathbf{H}^\top  \cdot \mathbf{\Lambda}^t_{\mathbf{z}} \cdot \mathbf{z}^t,
\end{matrix}
\label{eq:info_filter}
\end{align}
\note{(Author: a particular aspect of the implementation)}
\revise{In fact, due to the diagonal nature of the covariance matrix, the solution based on Eq.~\ref{eq:info_filter} can be greatly simplified, enabling efficient computation of the update. }Finally, the updated geometric state of gaussian surfels and corresponding covariance are computed by
\begin{align}
\mathbf{\hat{x}}^t = (\boldsymbol{\Lambda}^t)^{-1} \boldsymbol{\eta}^t,
\quad
\mathbf{\hat{{\Sigma}}}^t = (\boldsymbol{\Lambda}^t)^{-1}.
\end{align}
Ultimately, the updated position and normal from states $\textbf{x}^t$ are applied to the attributes $\textbf{p}_i$ and $\textbf{r}_i$ of surfel $S_i$. 
However, the estimation of relative rotation based on normal updates is underconstrained.
To address this issue, we introduce an additional rotational constraint based on the normal update, resulting in a well-defined 3D rotation. As illustrated in Fig.~\ref{fig:surfel_fusion}, we define \( \mathbf{n}_{tg} = \mathbf{n}_g \times \mathbf{n}_t \)
where \(\mathbf{n}_g\) is the original normal before update and \(\mathbf{n}_t\) is the updated normal. 
So the vector \( \mathbf{n}_{tg}\) represents the normal to the plane spanned by the two normals. 
The angle between them is \( \theta = \cos^{-1}(\mathbf{n}_g \cdot \mathbf{n}_t) \).
Based on this, we define a unique rotation represented by the rotation vector \( \Delta \textbf{R}(\textbf{n}_{tg}, \theta) \).
Therefore, for the rotation \(\mathbf{r}_i\) of surfel \(S_i\), whose matrix representation is \(\mathbf{R}_i\), the update is performed as follows:
\begin{align}
\hat{\textbf{R}}_i = \Delta \textbf{R}(\textbf{n}_{tg},\theta) \cdot \textbf{R}_i.
\label{eq:surfel_rotation_fusion}
\end{align}

\subsubsection{Differentiable Surfels Optimization}
\label{sec:surfel_optim}

After the explicit Gaussian surfel fusion based on upcoming measurements, we implement end-to-end optimization of the surfels using the rasterization-based rendering pipeline~\cite{kerbl3DGaussianSplatting2023}. 

\textbf{Reconstruction Loss}.
We utilize frame batches to optimize the local map through differentiable rendering, avoiding overfitting to the current frame. 
Specifically, we maintain a local map composed of the most recent \(N_{\text{batch}}\) frames.
After tracking a certain number of frames, we perform batch optimizations, where each iteration randomly selects a frame from the local map.
For thorough optimization, we perform a total of \(m \cdot N_{\text{batch}}\) iterations to ensure that each frame is iteratively optimized an average of \(m\) times.
The rendered color map, depth map, and normal map are constrained by the ground truth, resulting in the following loss terms:
\begin{align}   
  \mathcal{L}_c  =  \left |  C_k - \hat{C} _k \right |, \mathcal{L}_d  =  \left |  D_k - \hat{D} _k \right | ,  \mathcal{L}_n  = \left | 1 - \gamma \right | ,
\label{eq:loss_term}
\end{align}
where $\gamma = N_k \cdot \hat{N}_k $ represents the cosine of these two unit normal vector. Eq. (\ref{eq:loss_term}) utilizes image-space rendering to optimize Gaussian surfels through per-pixel supervision but lacks global consistency. Leveraging the geometric constraints of Gaussian surfels, we introduce per-surfel regularization constraints.

\textbf{Geometric Regularization}: 
We aim to ensure that the geometric properties of Gaussian surfels remain as consistent as possible during the current iterative optimization (avoiding deviations from the surface), while allowing for appearance changes introduced by new depth map. Therefore, we introduce the following geometric regularization term:
\begin{align}   
\mathcal{L}_{reg} =  \left | \textbf{p} - \textbf{p}_{f} \right | + w^n_{reg} \cdot\left |1 - \textbf{n}\cdot \textbf{n}_{f}   \right | ,
\label{eq:geo_term}
\end{align}
where $\textbf{p}_{f}$ and $\textbf{n}_{f}$ are derived from explicit Gaussian surfel fusion
and $w_{\text{reg}}^n$ is the weighting coefficient for the normal loss of the pre-surfel. Considering this per-surfel explicit geometric constraints in the end-to-end manner, the overall optimization loss is formulated as
\begin{align}   
  \mathcal{L}_{total} = \mathcal{L}_c + w_{d}\cdot\mathcal{L}_d + w_{n} \cdot \mathcal{L}_n + w_{reg}. \cdot \mathcal{L}_{reg}. 
  \label{eq:total_loss}
\end{align}
It is important to note that such regularization are not applicable to 3DGS-based methods~\cite{mono_gs,yanGSSLAMDenseVisual2024,splatam}. This is because 3DGS relies on a volumetric representation scheme that lacks explicit constraints with the scene geometry. Additionally, the inconsistency of inferred depths across multiple viewpoints contributes to the instability in its convergence.
Additionally, the proposed surfel initialization and fusion facilitate the state of the surfels to be optimized to be near convergence, thereby ensuring that this differentiable surfel optimization can achieve convergence performance of the map state within a few iterations.

\begin{figure*}[!h]
  \centering 
  \begin{minipage}{0.02\linewidth} 
    \centering
    \small
    \parbox[c][56.pt][c]{\textwidth}{\centering \rotatebox{90}{\texttt{Office2}}} \\
    \parbox[c][56.pt][c]{\textwidth}{\centering \rotatebox{90}{\texttt{Office3}}} \\
    \parbox[c][56.pt][c]{\textwidth}{\centering \rotatebox{90}{\texttt{Room1}}} \\
    \parbox[c][56.pt][c]{\textwidth}{\centering \rotatebox{90}{\texttt{Room2}}} \\
    \parbox[c][56.pt][c]{\textwidth}{\centering \rotatebox{90}{\texttt{Scene3}}} \\
    \parbox[c][56.pt][c]{\textwidth}{\centering \rotatebox{90}{\texttt{Scene4}}}
\end{minipage}%
\begin{minipage}{0.98\linewidth}
\includegraphics[width=1\linewidth]{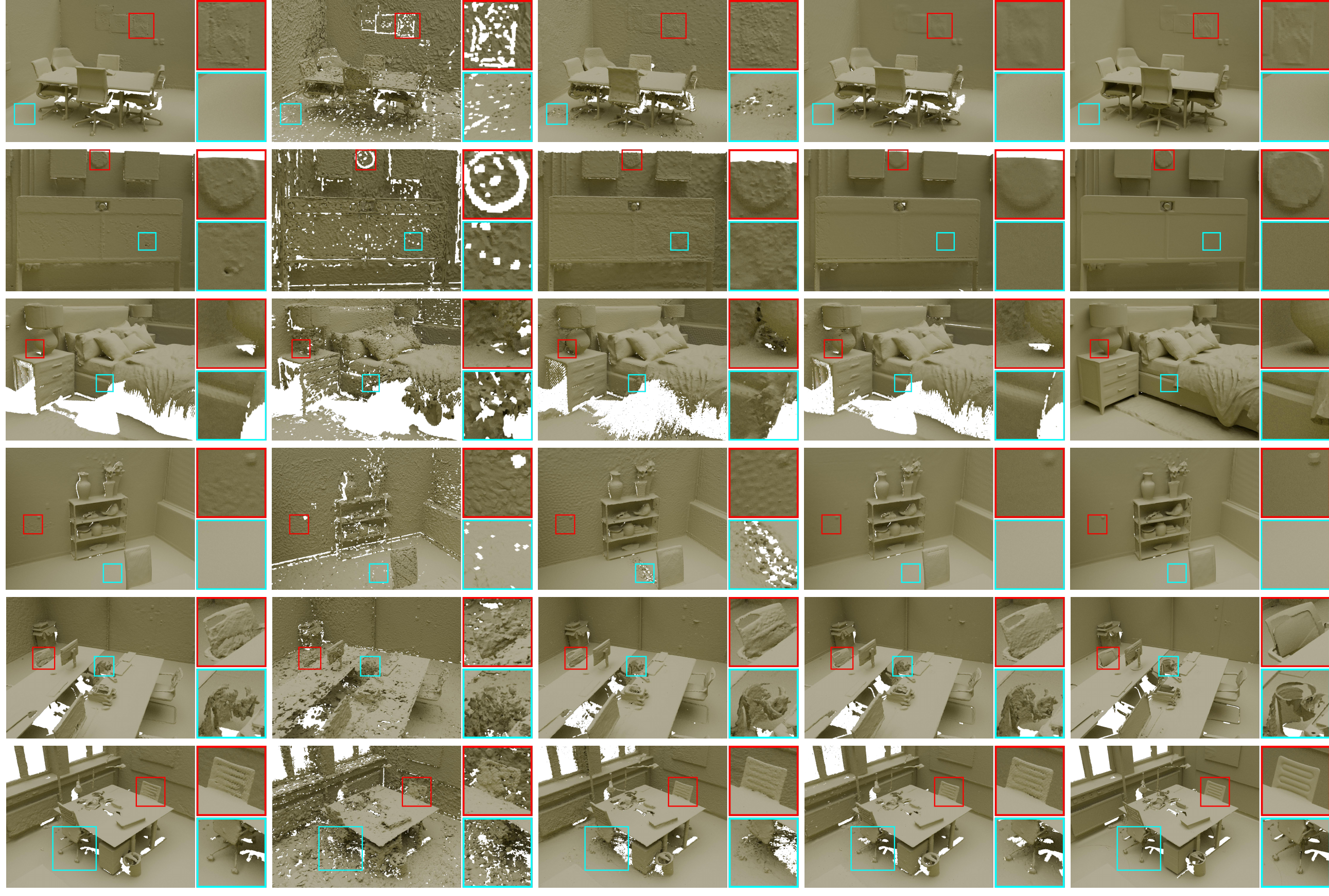}
\end{minipage} \\
\begin{minipage}{0.02\linewidth}
 \centering
 \fontsize{7pt}{8pt}\selectfont
 \textcolor{white}{x}
 \end{minipage}
 \hfill
 \begin{minipage}{0.2\linewidth}
 \centering
 \fontsize{7pt}{8pt}\selectfont
 Point-SLAM~\shortcite{point_slam}
 \end{minipage}
 \hfill
 \begin{minipage}{0.18\linewidth}
 \centering
 \fontsize{7pt}{8pt}\selectfont
 RTG-SLAM~\shortcite{rtg_slam}
 \end{minipage}
 \hfill
  \begin{minipage}{0.19\linewidth}
 \centering
 \small
 SplaTAM~\shortcite{splatam}
 \end{minipage}
 \hfill
 \begin{minipage}{0.19\linewidth}
 \centering
 \fontsize{7pt}{8pt}\selectfont
 Ours
 \end{minipage}
  \hfill
  \begin{minipage}{0.19\linewidth}
 \centering
 \fontsize{7pt}{8pt}\selectfont
 GT
 \end{minipage}
\caption{
  TSDF-based Reconstruction Result on Replica and ScanNet++. In terms of scene reconstruction mesh details on Replica~\cite{straubReplicaDatasetDigital2019} and ScanNet++~\cite{scannet++}, we outperform other methods with the overall quality and detail accuracy of the reconstructed mesh.
  }
  \label{fig:exp_recon_mesh}
\end{figure*}

\subsection{Camera Tracking}
\label{sec:pose_estim}

Given the current frame \( I_t \) and the global model \( \mathcal{S} \), camera tracking module aims to estimate the current camera pose \( \mathbf{T}_t \).
In optimization problems, we adopt the more compact and structure-preserving Lie algebra representation of camera poses as \( \boldsymbol{\xi}_t \in \mathfrak{se}(3) \).
For the robustness and accuracy of camera tracking, we propose a sparse-to-dense strategy.

\subsubsection{Sparse-Correspondence-based Pose Initialization}

Existing real-time reconstruction methods typically estimate camera poses through dense alignment between the current depth frame and global scene map. While achieving precise pose estimation, these methods are highly initial-value-dependent and prone to tracking failure during rapid camera motions.
To address this issue, we introduce a sparse feature correspondence-based pose estimation strategy for pose initialization. On one hand, it provides a stable initialization for the subsequent dense alignment; on the other hand, a good initialization significantly reduces the convergence time during the refinement stage.

Given a set of 2D-3D correspondences $\mathcal{M} = \{(\textbf{u}_i,\mathbf{X}_i^{w})\}_{i=0}^m$, where $\mathbf{u}_i \in \mathbb{R}^2$ denotes a 2D point in the current frame and $\mathbf{X}_i^w \in \mathbb{R}^3$ is the corresponding 3D point in the sparse map, we formulate the pose estimation as a reprojection error minimization problem. The initial camera pose $\boldsymbol{\xi}^{(0)}_t$ is estimated via Levenberg–Marquardt (LM) optimization:
\begin{align} 
  \boldsymbol{\xi}_t^{(0)} = \underset{\boldsymbol{\xi}_t}{\text{arg min}}  
\sum_{\mathcal{M}} \rho\left ( 
\left | \mathbf{u}_i - \Pi\left ( \exp(\boldsymbol{\xi}_t)\cdot \textbf{X}_i^w \right )
 \right | ^2
  \right ) ,
\end{align}
Here, $\exp(\cdot)$ is the exponential map from the Lie algebra to the Lie group and $\rho(\cdot)$ is a robust loss function used to mitigate the influence of outliers. A similar approach is also adopted in~\cite{huang2024photo}

{\small 
   \begin{table}
       \caption{Tracking performance (ATE RMSE$[cm]$) on Replica.}
       \centering
       \tabcolsep=0.15cm
       \renewcommand\arraystretch{1.05}
       \resizebox{\columnwidth}{!}{
       \begin{tabular}{l c c c c c c c c c}
       \toprule[1pt]
           Method     & \texttt{Rm0}  & \texttt{Rm1} & \texttt{Rm2} & \texttt{Off0} & \texttt{Off1} & \texttt{Off2} & \texttt{Off3} & \texttt{Off4} & Avg. \\ \hline
           MASt3R~\shortcite{leroy2024mast3r}
           & 1.07 & 0.99 & 0.87 & 0.90 & 4.90 & 1.21 & 1.77 & 1.63 & 1.67 \\
           SLAM3R~\shortcite{liu2024slam3r}
           & 4.56 & 5.88 & 5.72 & 11.17 & 6.32 & 6.15 & 4.95 &  8.09 & 6.61\\
           \hline    

           NICE-SLAM~\shortcite{nice_slam} 
           & 0.97 & 1.31 & 1.07 & 0.88 & 1.00 & 1.06 & 1.10 & 1.13 & 1.06 \\
           Vox-Fusion~\shortcite{vox_fusion}
           & \cellcolor{yellow!40}0.41 & 0.50 & 0.52 & 0.47 & 0.61 & 0.67 & 0.47 & \cellcolor{yellow!40}0.47 & 0.52 \\
           Point-SLAM~\shortcite{point_slam}
           & 0.54 & 0.43 & 0.34 & \cellcolor{yellow!40}0.36 & 0.45	& 0.44 & 0.63 & 0.72 & 0.49  \\
           SplaTAM~\shortcite{splatam}
           & 0.47 &\cellcolor{yellow!40} 0.42 & \cellcolor{yellow!40}0.32& 0.46 & \cellcolor{yellow!40}0.24 & \cellcolor{yellow!40}0.28 & \cellcolor{yellow!40}0.39 & 0.56 & \cellcolor{yellow!40}0.39 \\
           RTG-SLAM~\shortcite{rtg_slam}
           & \cellcolor{orange!40}0.20 & \cellcolor{red!40}0.18 & \cellcolor{orange!40}0.13 & \cellcolor{orange!40}0.22 & \cellcolor{red!40}0.12 & \cellcolor{orange!40}0.22 & \cellcolor{orange!40}0.20 & \cellcolor{red!40}0.19 & \cellcolor{orange!40}0.18 \\
           \textbf{Ours} 
           & \cellcolor{red!40}0.18 & \cellcolor{red!40}0.18 & \cellcolor{red!40}0.11 & \cellcolor{red!40}0.15	& \cellcolor{red!40}0.12 & \cellcolor{red!40}0.19 & \cellcolor{red!40}0.17 & \cellcolor{orange!40}0.20 & \cellcolor{red!40}0.17 \\
           
       \bottomrule[1pt]
       \end{tabular}
       }
       \label{tab:replica_tracking}
   \end{table} 
}

\subsubsection{Refine with Dense Alignment}
After obtaining the initial pose, we perform dense alignment for pose refinement, which formulates a nonlinear least-squares problem to jointly optimize the camera pose based on geometric and photometric measurements.
Consistent with previous work~\cite{whelanElasticFusionDenseSLAM2015c, daiBundleFusionRealtimeGlobally2017c}, we align the vertex map of the current frame \( V_t \) with the vertex map \( V_{\mathcal{G}} \) and normal map \( N_{\mathcal{G}} \) of the global model to construct the geometric error term, where $(\cdot)_{\mathcal{G}}$ denote the global model:
\begin{align} 
E(\boldsymbol{\xi}_t)_{\text{icp}} =
\sum_{\mathcal{N}}
\left |
\left (V_{\mathcal{G}}\left ( \textbf{u}'_i \right )  - \exp{ \left ( \boldsymbol{\xi}_{t} \right) }V_t\left (\textbf{u}_i\right )  \right )  \cdot  N_\mathcal{G}\left (\textbf{u}'_i\right)
\right | _2 ^2,
\end{align}
Where \( \mathbf{u}'_i = \Pi\left ( \exp(\boldsymbol{\xi}) \cdot \Pi^{-1}(\textbf{u}_i) \cdot d_i \right ) \), and \( \mathcal{N} \) is the set of geometrically matched points after filtering.
Subsequently, by comparing the rendered image with the RGB observation, the photometric error is defined as:
\begin{align}   
E(\boldsymbol{\xi}_t)_{\text{photo}} =  \sum_{\mathcal{N}} \left | C_\mathcal{G}\left ( \textbf{u}'_i \right ) - C_t\left ( \textbf{u}_i \right )  \right | _2 ^2,
\end{align}
Finally, we adopt a joint optimization strategy to minimize:
\begin{align}   
E(\boldsymbol{\xi}_t)_{\text{dense}} =  E_{\text{icp}} + \lambda_{\text{photo}}\cdot E_{\text{photo}}.
\end{align}
We use $\lambda_{\text{photo}}$ to balance the influence of photometric constraints on pose optimization.
To mitigate potential degeneration in extreme cases, we further evaluate the convergence behavior of the dense alignment stage to determine whether its result should be adopted.

\section{Experiments}
\label{sec:experiments}

\subsection{Experimental Setup}
\label{sec:exp_setup}
\noindent
\textbf{Datasets.}
We conduct experiments on Replica~\cite{straubReplicaDatasetDigital2019}, TUM-RGBD~\cite{sturmBenchmarkEvaluationRGBSLAM2012} and ScanNet++~\cite{scannet++}.
Replica provides camera trajectories and scene depth, which we use to assess localization accuracy, reconstruction quality, and rendering performance. 
TUM-RGBD offers ground truth trajectories captured by external motion capture devices, serving as the benchmark for evaluating tracking accuracy.
ScanNet++ provides high-quality images and accurate depth maps, which we use to evaluate reconstruction and rendering quality.
For the selection of evaluation sequences, we maintain consistency with previous works~\cite{point_slam, rtg_slam}.
Furthermore, we captured three real-world sequences using the consumer-grade Azure Kinect sensor, enabling qualitative evaluation and comparison. 
\note{(Author: add the detail of the captured Azure RGBD dataset)}
\revise{This dataset contains three scenes: \textit{Bag}, \textit{Hydrant}, and \textit{Stone}, as shown in Fig.~\ref{fig:teaser}. The motion pattern involves slow movements around objects within the scenes. The challenges include: 1) missing depth data in certain regions (due to exceeding the sensor range or measurement failures caused by surface reflection and transmission), 2) realistic scene lighting, and 3) continuous orbiting motion.}

{\small 
    \begin{table}
        \caption{Tracking performance (ATE RMSE$[cm]$) on TUM-RGBD. \ding{55} represents tracking failure. 
        $^{*}$ indicates that the result is taken from the original paper.
        $^{\dagger}$ denotes the offline variants.}
        \centering
        \tabcolsep=0.15cm
        \renewcommand\arraystretch{1.05}
        \resizebox{\columnwidth}{!}{
        \begin{tabular}{@{}c |l c c c c c c}
        \toprule[1pt]
            & Method & \texttt{fr1/desk} & \texttt{fr1/desk2} & \texttt{fr1/room}  & \texttt{fr2/xyz} & \texttt{fr3/office} & Avg.   \\ 

            \hline
            \multirow{3}{*}{\rotatebox{90}{\textit{Classical} }}
            &ElasticFusion~\shortcite{whelanElasticFusionDenseSLAM2015c}
            & 2.53 
            & 6.83 
            & 21.49 
            & 1.17 
            & 2.52 
            & 6.91 \\
            &ORB-SLAM2~\shortcite{mur2017orb} 
            &2.16	
            &2.99	
            &16.14	
            &0.50	
            &1.56	
            &4.68 \\
            &BAD-SLAM~\shortcite{schopsBADSLAMBundle2019a}
            & 1.70 
            & \ding{55} 
            & \ding{55} 
            & 1.10 
            & 1.70 
            & \ding{55} \\
            
            \hline
            \multirow{6}{*}{\rotatebox{90}{\textit{Differentiable}}}
            &NICE-SLAM$^{*}$~\shortcite{nice_slam}
            & 4.26 & 4.99 & 34.49 &  31.73 & 3.87 & 15.87  \\
            &Vox-Fusion~\shortcite{vox_fusion}
            & \cellcolor{yellow!40}2.54 
            & \cellcolor{yellow!40}3.95
            & \cellcolor{orange!40}12.51 
            & 1.40 
            & 26.01
            & 9.28  \\
            &Point-SLAM$^{*}$~\shortcite{point_slam}
            & 4.34 
            & 4.54 
            & 30.92 
            & 1.31 
            & \cellcolor{yellow!40}3.48 
            & 8.92 \\
            &SplaTAM$^{*}$~\shortcite{splatam}
            & 3.35 
            & 6.54 
            & \cellcolor{red!40}11.13 
            & \cellcolor{yellow!40}1.24 
            & 5.16 
            & \cellcolor{yellow!40}5.48  \\
            &RTG-SLAM~\shortcite{rtg_slam}
            &\cellcolor{orange!40}2.30	
            &\cellcolor{red!40}2.77	
            &17.43
            &\cellcolor{orange!40}1.15
            &\cellcolor{red!40}1.41	
            &\cellcolor{orange!40}5.12 \\
            &\textbf{Ours}
            & \cellcolor{red!40}2.21 
            & \cellcolor{orange!40}3.09 
            & \cellcolor{yellow!40}14.68
            & \cellcolor{red!40}0.98 
            & \cellcolor{red!40}1.41 
            & \cellcolor{red!40}4.47

            \\ \hline\hline
            \multirow{3}{*}{\rotatebox{90}{\textit{offline} }}
            
            &\cellcolor{gray!10}ORB-SLAM2$^{\dagger}$~\shortcite{mur2017orb}
            & \cellcolor{gray!10}1.53 
            & \cellcolor{gray!10}2.21 
            & \cellcolor{gray!10}5.38 
            & \cellcolor{gray!10}0.40 
            & \cellcolor{gray!10}0.91 
            & \cellcolor{gray!10}2.09 \\
            &\cellcolor{gray!10}{RTG-SLAM$^{\dagger}$~\shortcite{rtg_slam}}
            & \cellcolor{gray!10}1.61
            & \cellcolor{gray!10}2.61
            & \cellcolor{gray!10}4.18
            & \cellcolor{gray!10}0.38
            & \cellcolor{gray!10}1.31
            & \cellcolor{gray!10}2.02   \\
            &\cellcolor{gray!10}\textbf{Ours}$^{\dagger}$
            &\cellcolor{gray!10}1.56	
            &\cellcolor{gray!10}2.45	
            &\cellcolor{gray!10}4.34	
            &\cellcolor{gray!10}0.43	
            &\cellcolor{gray!10}1.13	
            &\cellcolor{gray!10}1.98 \\

        \bottomrule[1pt]
        \end{tabular}
        }
        \label{tab:tum_tracking}
    \end{table} 
}

\noindent
\textbf{Metric.}
To evaluate the accuracy of camera tracking, we use ATE RMSE ~\cite{sturm2012benchmark} as the metric. 
For reconstruction quality, we adopt the following metrics: accuracy, accuracy ratio [\textless3\textit{cm}], completeness, and completeness ratio [\textless3\textit{cm}], following~\cite{nice_slam, rtg_slam}.
Regarding rendering performance, we generate full-resolution rendered images and utilize three metrics for evaluation: PSNR~\cite{hore2010image}, SSIM~\cite{wang2004image}, and LPIPS~\cite{zhang2018unreasonable}. 

\begin{figure*}[!h]
\includegraphics[clip,width=0.98\linewidth, trim={0.0cm 8.5cm 0cm 0cm}]{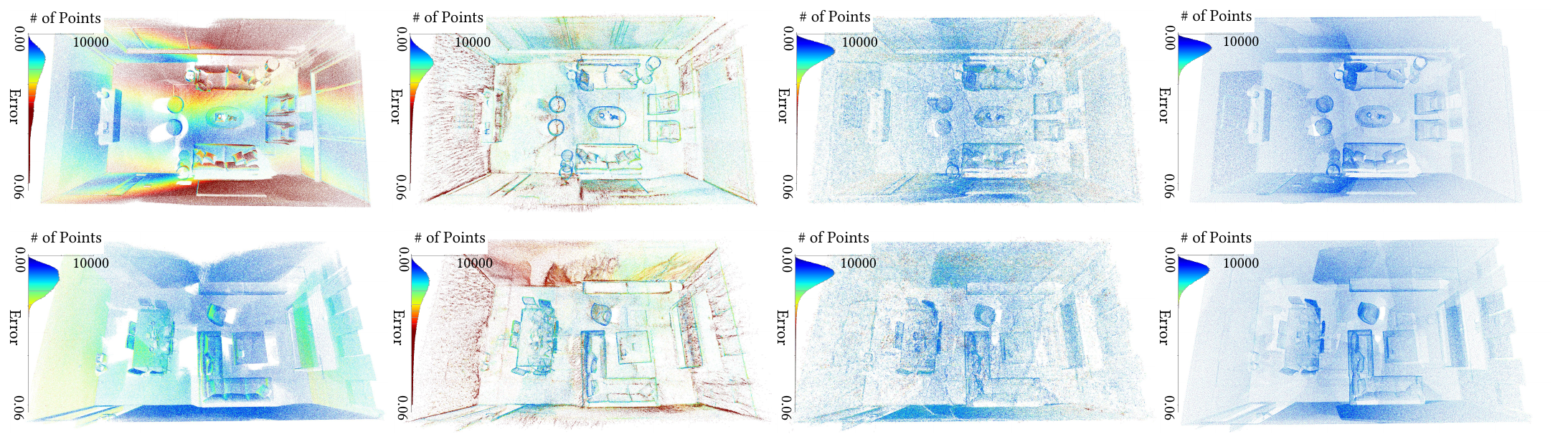} \\

\begin{minipage}{0.242\linewidth}
\centering
\fontsize{7pt}{8pt}\selectfont
ElasticFusion~\shortcite{whelanElasticFusionDenseSLAM2015c}
\end{minipage}
\hfill
\begin{minipage}{0.242\linewidth}
\centering
\fontsize{7pt}{8pt}\selectfont
SplaTAM~\shortcite{splatam}
\end{minipage}
\hfill
\begin{minipage}{0.242\linewidth}
\centering
\fontsize{7pt}{8pt}\selectfont
RTG-SLAM~\shortcite{rtg_slam}
\end{minipage}
\hfill
\begin{minipage}{0.242\linewidth}
\centering
\fontsize{7pt}{8pt}\selectfont
Ours
\end{minipage}
 
  \caption{
  Geometry accuracy of points. Our method achieved globally high-precision reconstruction. In contrast, other methods exhibited higher errors in either the details of local geometric complexity (RTG-SLAM~\cite{rtg_slam}) or the overall geometric structure (SplaTAM~\cite{splatam} and ElasticFusion~\cite{whelanElasticFusionDenseSLAM2015c}).
  }
  \label{fig:exp_recon_pc}
\end{figure*}

\noindent
\textbf{Baseline.}
The compared methods include
differentiable rendering-based SLAM: NICE-SLAM~\cite{nice_slam}, Vox-Fusion~\cite{vox_fusion}, Point-SLAM~\cite{point_slam}, SplaTAM~\cite{splatam}, and RTG-SLAM~\cite{rtg_slam}, 
and dense visual SLAM:
ElasticFusion~\cite{whelanElasticFusionDenseSLAM2015c}, ORB-SLAM2~\cite{mur2017orb}, and BAD-SLAM~\cite{schopsBADSLAMBundle2019a}. 
At the same time, we also include partial comparisons with the latest feed-forward model-based methods including MASt3R~\cite{leroy2024mast3r} and SLAM3R~\cite{liu2024slam3r}.
All these methods provide open-source implementations for reproducibility and comparison.

\subsection{Main Experiments}
\noindent\textbf{Evaluation of Tracking.}
The tracking accuracy results on Replica dataset are shown in Tab.~\ref{tab:replica_tracking}. 
All the results of other methods except Vox-Fusion are from~\cite{rtg_slam}.
Compared to other methods, our method achieves the highest average accuracy and attains the best trajectory precision in most scenarios. 
The superiority of the proposed method in tracking accuracy stems from its sparse-to-dense tracking strategy, as well as the geometry-aware differentiable scene representation.
Tab.~\ref{tab:tum_tracking} presents the tracking accuracy comparison of different methods on the TUM-RGBD dataset. We categorize the existing approaches into three groups: (1) \textit{Classical} methods, (2) \textit{Differentiable} scene representation based methods, and (3) \textit{Offline} variants.
For the \textit{Classical} methods, we adopted the results from~\cite{nice_slam, point_slam}, which have also been widely used by the following works~\cite{vox_fusion, mono_gs, splatam, rtg_slam}.
The inclusion of \textit{offline} variants is motivated by the fact that the official implementations of both RTG-SLAM and ORB-SLAM2 produce globally optimized trajectories (\textit{i.e.}, in offline mode). 
To ensure fair comparison, we recorded their online output while also developing an offline version of EGG-Fusion as reference, 
\note{(Author: the offline version)}
\revise{which is implemented similarly to RTG-SLAM’s official release, both of which are based on the ORB-SLAM2 backend. After tracking, both perform a global optimization and then output the final trajectory.}
Experimental results demonstrate that our method achieves the highest average tracking accuracy compared to all real-time SLAM systems in real-world scenarios.

\noindent\textbf{Evaluation of Reconstruction.}
Different systems employ diverse scene representations, leading to significant variations in the format of reconstruction results. 
For a fair comparison across different methods, we adopt two evaluation schemes following the approach in~\cite{rtg_slam, point_slam}.
For explicit reconstruction results, we uniformly sample a fixed number ($1\times 10^{6}$) of points from the final reconstructed Gaussian primitives for evaluation. 
For implicit reconstruction result, we utilize depth maps with ground truth camera poses to perform TSDF fusion, followed by mesh extraction via the marching cubes to assess the final reconstruction accuracy~\cite{nice_slam}.
The reconstruction accuracy of different methods on Replica and ScanNet++ is shown in Tab.~\ref{tab:reconstruction}.
Under both evaluation methods, our reconstruction accuracy surpasses that of other methods, with the exception of Point-SLAM.  
It is noteworthy that the high precision of Point-SLAM from its reliance on ground truth depth to determine rendering sampling positions, which grants it an unfair advantage in depth evaluation.
In both datasets, our average Acc. Ratio approaches 100\%, indicating that nearly all reconstructed surfels lie within 3\textit{cm} of the true scene surface. 
The qualitative comparison of reconstruction results is shown in Fig.~\ref{fig:exp_recon_mesh} and Fig.~\ref{fig:exp_recon_pc}.

\begin{figure*}[t]
  \centering 
\includegraphics[clip,width=0.98\linewidth, trim={0.0cm 35.5cm 0cm 0cm}]{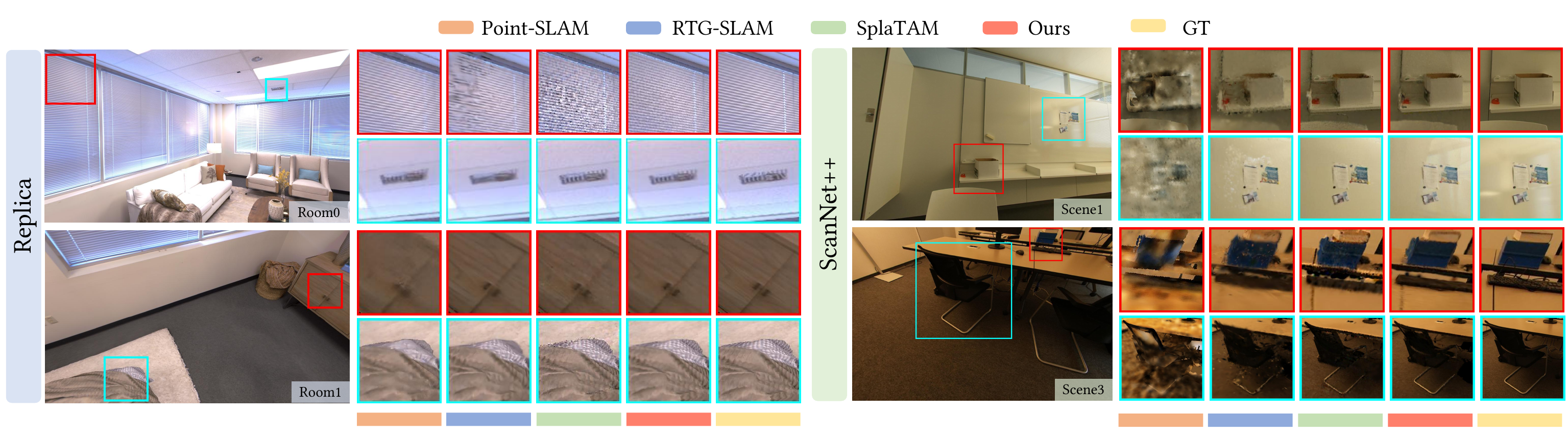}
\caption{
 Rendering Results on Replica and ScanNet++. We present a comparison of novel view synthesis results on Replica~\cite{straubReplicaDatasetDigital2019} and ScanNet++~\cite{scannet++}. Our method demonstrates superior rendering details in both training views (Replica, left) and testing views (ScanNet++, right).
  }
  \label{fig:exp_render}
\end{figure*}

{
\begin{table}
    \caption{Reconstruction performance on Replica/ScanNet++. N/A indicates missing data. $^*$ indicates that the result is taken from the original paper or~\cite{rtg_slam}.} 
    \centering
    \tabcolsep=0.07cm
    \renewcommand\arraystretch{1.05}
    \resizebox{\columnwidth}{!}{

    \tiny
    \begin{tabular}{@{}c | l c c c c }
    \toprule[0.6pt]
        \texttt{}
        & Method & Acc.$\downarrow$ & Acc. Ratio$\uparrow$ & Comp.$\downarrow$ & Comp. Ratio$\uparrow$ \\
        
        \hline

        & MASt3R$^{*}$~\shortcite{leroy2024mast3r}
        & 4.71 / N/A 
        & 78.12 / N/A
        & 3.36 / N/A
        & 79.32 / N/A \\
        
        & SLAM3R$^{*}$~\shortcite{liu2024slam3r}
        & 3.57 / N/A 
        & 82.23 / N/A
        & 2.62 / N/A
        & 81.29 / N/A\\
        \cline{2-6}
        
        \multirow{6}{*}{\rotatebox{90}{\tiny TSDF}} 
        &{NICE-SLAM$^{*}$~\shortcite{nice_slam}}
        & 2.84/4.45 
        & 84.44/74.49 
        & \fcolorbox{white}{orange!40}{2.31}/2.04 
        &  84.97/86.63 \\ 
        &{Vox-Fusion~\shortcite{vox_fusion}}
        & 2.03/N/A  
        & 89.04/N/A 
        & 2.58/N/A 
        & 86.93/N/A \\ 
        &{Point-SLAM~\shortcite{point_slam}} 
        & \fcolorbox{white}{red!40}{0.76}/\fcolorbox{white}{red!40}{0.67} 
        & \fcolorbox{white}{red!40}{99.80}/\fcolorbox{white}{red!40}{99.12} 
        & 2.42/\fcolorbox{white}{red!40}{0.68} 
        & \fcolorbox{white}{yellow!40}{87.46}/\fcolorbox{white}{orange!40}{98.94} \\ 
        &{SplaTAM~\shortcite{splatam}}       
        & \fcolorbox{white}{yellow!40}{1.12}/\fcolorbox{white}{yellow!40}{1.70 }
        & \fcolorbox{white}{yellow!40}{96.31}/\fcolorbox{white}{yellow!40}{90.87 }
        & 2.49/\fcolorbox{white}{yellow!40}{0.83}
        & \fcolorbox{white}{orange!40}{87.52}/98.51 \\
        &{RTG-SLAM~\shortcite{rtg_slam}}
        & 1.30/1.86 
        & 91.96/82.09
        & \fcolorbox{white}{yellow!40}{2.37}/1.00 
        & 86.81/\fcolorbox{white}{yellow!40}{98.72} \\

        &\textbf{Ours}
        & \fcolorbox{white}{orange!40}{0.90}/\fcolorbox{white}{orange!40}{1.51} 
        & \fcolorbox{white}{orange!40}{97.60}/\fcolorbox{white}{orange!40}{94.20} 
        & \fcolorbox{white}{red!40}{2.25}/\fcolorbox{white}{orange!40}{0.76} 
        & \fcolorbox{white}{red!40}{87.73}/\fcolorbox{white}{red!40}{99.20} \\
        \hline
        
        \multirow{4}{*}{\rotatebox{90}{\tiny Points}} 
        &{ElasticFusion\shortcite{whelanElasticFusionDenseSLAM2015c}} 
        & 1.38/N/A 
        & 91.72/N/A 
        & 7.38/N/A 
        & 65.50/N/A \\
        &{SplaTAM~\shortcite{splatam}} 
        & 2.87/1.71 
        & 74.27/89.15 
        & 3.58/1.60 
        & 71.72/92.64 \\
        &{RTG-SLAM~\shortcite{rtg_slam}} 
        & 0.80/1.06
        & 98.52/95.34
        & 2.88/1.22
        & 81.92/95.78 \\
        &\textbf{Ours}
        &\fcolorbox{white}{red!40}{0.60}/\fcolorbox{white}{red!40}{0.67}
        &\fcolorbox{white}{red!40}{99.99}/\fcolorbox{white}{red!40}{99.98}
        &\fcolorbox{white}{red!40}{2.67}/\fcolorbox{white}{red!40}{0.91}
        &\fcolorbox{white}{red!40}{84.88}/\fcolorbox{white}{red!40}{99.04} \\
    \bottomrule[0.6pt]
    \end{tabular}
    }
    \label{tab:reconstruction}
\end{table} 
}

\begin{table}[t]
\centering
\scriptsize
\tabcolsep=0.2cm
\renewcommand\arraystretch{1.05}

\caption{Rendering performance on ScanNet++.}
\label{tab:scannetpp_part}

\begin{tabular}{l ccc | ccc}
\toprule

\multirow{2}{*}{Methods} & \multicolumn{3}{c}{Novel View} & \multicolumn{3}{c}{Training View} \\

 & PSNR $\uparrow$ & SSIM $\uparrow$ & LPIPS $\downarrow$ & PSNR $\uparrow$ & SSIM $\uparrow$ & LPIPS $\downarrow$\\

\midrule

Point-SLAM~\shortcite{point_slam}
 & 17.68 & 0.623 & 0.548 & 24.35 & 0.800 & 0.373 \\

SplaTAM~\shortcite{splatam}
 & \cellcolor{orange!40}24.75 & 0.900 & \cellcolor{red!40}0.208 & 27.30 & \cellcolor{orange!40}0.940 & \cellcolor{red!40}0.130 \\

RTG-SLAM~\shortcite{rtg_slam}
& 24.77 & \cellcolor{orange!40}0.882 & 0.255 & \cellcolor{orange!40}27.54 & 0.925 & 0.184 \\

\textbf{Ours}
& \cellcolor{red!40}25.70 & \cellcolor{red!40}0.907 & \cellcolor{orange!40}0.212 & \cellcolor{red!40}29.06 & \cellcolor{red!40}0.944 & \cellcolor{orange!40}0.141\\

\bottomrule
\end{tabular}
\label{tab:scannetpp_rendering}
\end{table}

\noindent\textbf{Evaluation of Novel View Synthesis.}
We report the rendering results on the ScanNet++ dataset. 
As shown in Tab.~\ref{tab:scannetpp_rendering}, we achieve the best rendering quality on both the training views and unseen test views. 
This demonstrates the generalization capability of our Gaussian surfel based scene representation for novel view synthesis.
This success can be attributed to the proposed method's ability to tightly adhere Gaussian surfels to the scene surface, ensuring excellent geometric consistency across different viewpoints.
The qualitative comparison of rendering results is shown in the Fig.~\ref{fig:exp_render}.

\noindent\textbf{System Performance.}
We evaluated and reported the system performance of several methods. 
To ensure a relatively fair comparison, we used the default configuration for each method and conducted time and memory consumption statistics on the Replica \texttt{office0}. Specifically, we measured the time taken for each frame in the main tracking and mapping modules, reporting the average values. We also recorded the number of iterations in tracking and mapping, as well as the average time per iteration. 
Then we can derive the theoretical FPS for the system under a single-threaded mode.
As shown in the Tab.~\ref{tab:system_performance}, our method outperforms the current SOTA methods in tracking, mapping and overall system FPS. 
It is worth noting that, thanks to the carefully designed system, scene primitives converge rapidly with minimal updates per optimization, resulting in significantly lower average mapping time per frame compared to other methods.
Moreover, our online memory consumption is also significantly lower than that of other approaches.

{\footnotesize 
    \begin{table}
        \caption{
        Comparison of time and memory performance on Replica \texttt{Off0}. $^{*}$ denote the average cost time of sparse correspondence based pose estimation and dense alignment, respectively.
        }
        \centering
        \tabcolsep=0.15cm
        \resizebox{\columnwidth}{!}{
        \renewcommand\arraystretch{1.2}
        \begin{tabular}{l l c c c c c}
        \toprule[1pt]
            Method  
            & \makecell{Tracking\\$[\text{ms} \times \text{it}]$$\downarrow$} 
            & \makecell{Mapping\\$[\text{ms} \times \text{it}]$$\downarrow$} 
            & \makecell{Mapping\\/Frame[s]$\downarrow$} 
            & \makecell{Model Size\\$[\text{MB}]$$\downarrow$}
            & \makecell{Mem.\\$[\text{GB}]$$\downarrow$}
            & \makecell{FPS}$\uparrow$\\ \hline
            NICE-SLAM\shortcite{nice_slam}
            & 6.6 $\times$ 10 
            & 28.6$\times$ 60 
            & 1.717 
            & \cellcolor{orange!40}48.48
            & 9.8
            & \cellcolor{yellow!40}2.91 \\
            Vox-Fusion\shortcite{vox_fusion}
            & 16.5 $\times$ 30 
            & \cellcolor{yellow!40}34.8 $\times$ 15 
            & \cellcolor{yellow!40}0.675 
            & \cellcolor{red!40}1.49
            & \cellcolor{yellow!40}7.8
            & 0.75 \\
            Point-SLAM\shortcite{point_slam}
            & \cellcolor{yellow!40}10.1 $\times$ 40 
            & 31.18 $\times$ 300 
            & 9.892 
            & 15431
            & 9.8
            &  0.40 \\
            SplaTAM\shortcite{splatam}
            & 45.2 $\times$ 40 
            & 54.7 $\times$ 60 
            & 3.283 
            & 310
            & 9.1
            & 0.19 \\
            RTG-SLAM\shortcite{rtg_slam}   
            & \cellcolor{orange!40}29.1 $\times$ 1 
            & \cellcolor{orange!40}3.5 $\times$ 50  
            & \cellcolor{orange!40}0.207 
            & \cellcolor{yellow!40}51
            & \cellcolor{orange!40}2.7
            & \cellcolor{orange!40}15.73 \\
            \textbf{Ours}
            & \cellcolor{red!40}(14.2 + 9.4)$^{*}$ 
            & \cellcolor{red!40}7.5 $\times$ 9 
            & \cellcolor{red!40}0.071 
            & 150
            & \cellcolor{red!40}1.8
            & \cellcolor{red!40}24.21 \\
        \bottomrule[1pt]
        \end{tabular}
        }
        \label{tab:system_performance}
    \end{table} 
}

{\small 
    \begin{table}
        \caption{Ablation study of sparse-to-dense tracking.}
        \centering
        \tabcolsep=0.1cm
       \renewcommand\arraystretch{1.05}
        \resizebox{\columnwidth}{!}{
        \begin{tabular}{l c c c c c c}
        \toprule[1pt]
            Method     & \texttt{fr1/desk} & \texttt{fr1/desk2} & \texttt{fr1/room}  & \texttt{fr2/xyz} & \texttt{fr3/office} & Avg.   \\ \hline
            \textit{w/o. sparse}
            & 6.39 & 9.29 & \ding{55}  & 1.43 & \ding{55}  & \ding{55}  \\
            $ \textit{full}$
            &\textbf{2.21}	&\textbf{3.09}	&\textbf{14.68}	&\textbf{0.98}	&\textbf{1.41}	&\textbf{4.47}\\
        \bottomrule[1pt]
        \end{tabular}
        }
        \label{tab:tracking_ablation}
    \end{table} 
}

\begin{table}[t]
\centering
\scriptsize
\tabcolsep=0.25cm
\renewcommand\arraystretch{1.1}

\caption{\revise{Ablation study of surfel fusion on ScanNet++.}}
\begin{tabular}{l cc | cc}
\toprule
\multirow{2}{*}{Metric} & \multicolumn{2}{c}{TSDF} & \multicolumn{2}{c}{Points} \\
 & \textit{w/o fusion} & \textit{full} & \textit{w/o fusion} & \textit{full} \\
\midrule
ACC./Acc. Ratio & 1.59 / 93.42 & \textbf{1.51} /  \textbf{94.20} & 0.73 / 99.28 & \textbf{0.67} / \textbf{99.98} \\
Comp./Comp. Ratio & 0.80 / 98.73 & \textbf{0.76} / \textbf{99.20} & 1.01 / 98.93 & \textbf{0.91} / \textbf{99.04} \\
 
\bottomrule
\end{tabular}
\label{tab:recon_ablation}
\end{table} 

\noindent\textbf{Qualitative Comparison.}
\begin{figure*}[!h]
  \centering 
\begin{minipage}{\linewidth}
\includegraphics[clip,width=1.0\linewidth, trim={0cm 7.5cm 0cm 0cm}]{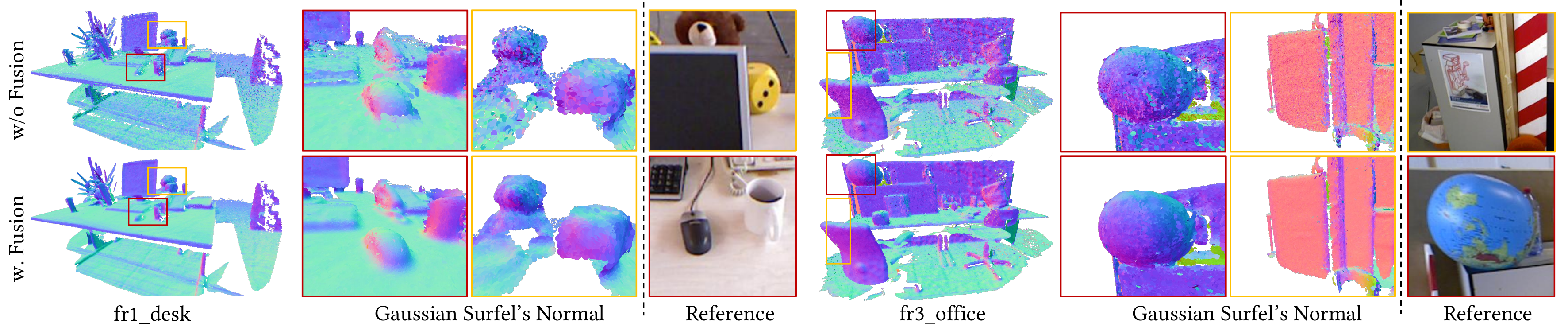}
 \end{minipage} 
  \caption{
Ablation of Surfel Fusion with Information Filter.
The strategy of surfel fusion with information filter demonstrates robustness against noise, effectively enhancing the accuracy of surface reconstruction on the both scenes (\texttt{fr1/desk}, \texttt{fr3/office}) in TUM-RGBD~\cite{sturm2012benchmark} dataset.
}
  \label{fig:exp_surfel_fusion_ablation}
\end{figure*}
As shown in Fig.~\ref{fig:teaser}, we collected three outdoor object-centric RGB-D sequences using Azure Kinect for qualitative demonstration of our method's performance and comparison with other approaches. Our method enables real-time, online, high-quality object-level reconstruction, while also generating a surface representation based on Gaussian surfels with confidence. 
Additionally, we render the scene from novel viewpoints. Even under large viewpoint changes compared to the training views, our geometry-aware Gaussian surfel representation achieves significantly higher rendering quality than other methods.

\subsection{Ablation Study}

\begin{figure}[!h]
  \centering 
  \includegraphics[clip,width=0.98\linewidth, trim={0.0cm 10.cm 4.5cm 0cm}]{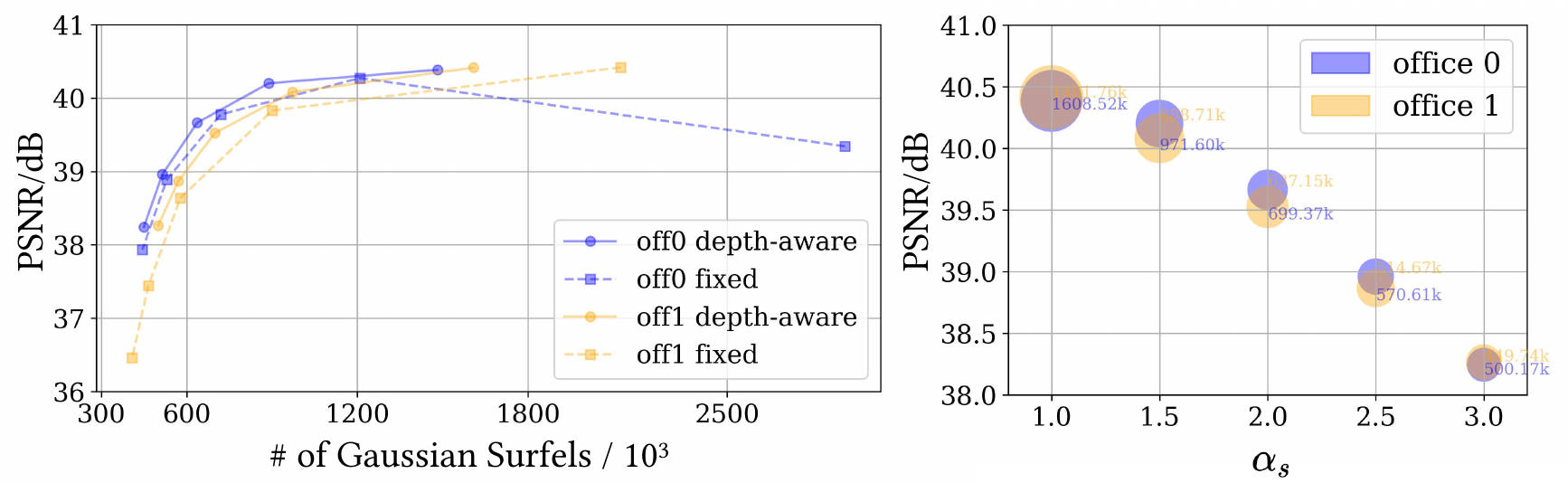}
  \caption{Depth-aware vs. Fixed scale initialization. The relationship between the number of surfels and rendering quality under different scale initialization strategies (left). the impact of different scale scaling factors on the number of surfels and rendering quality (right). }
  \label{fig:exp_surfel_scale_ablation}
\end{figure}
\begin{figure}[!h]
  \centering 
\begin{minipage}{\linewidth}
\includegraphics[clip,width=1.0\linewidth, trim={0cm 10.cm 10cm 0cm}]{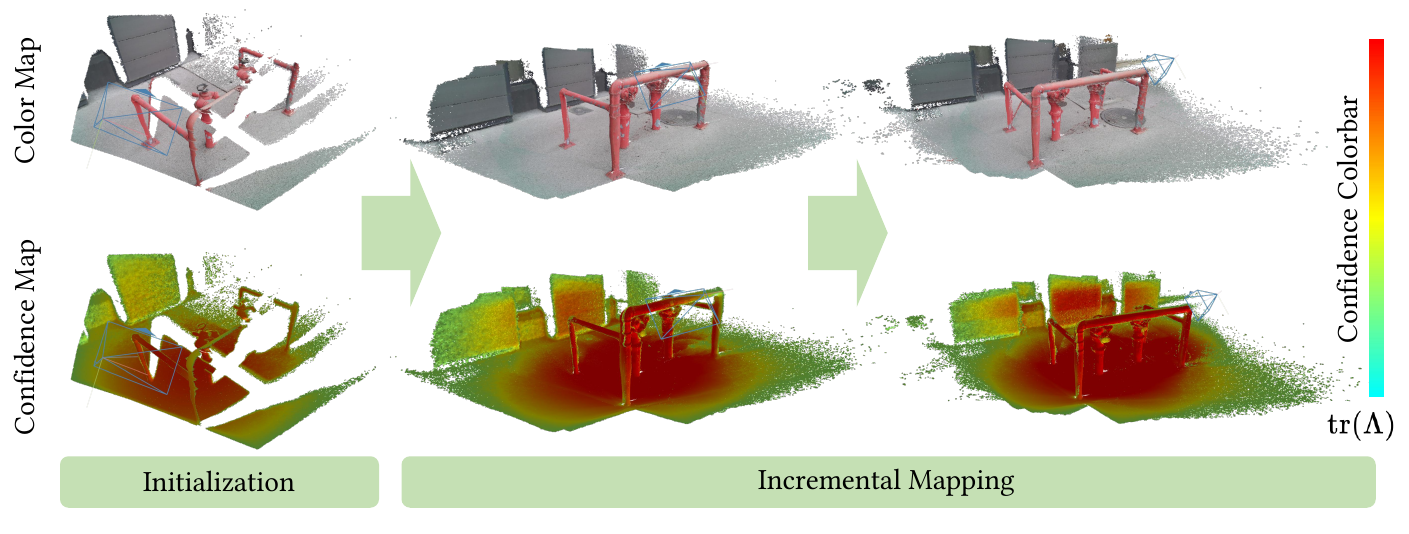}
 \end{minipage} 
  \caption{
Confidence Update with Incremental Mapping. 
We use the trace of information matrix $\text{tr}(\Lambda)$ as the surfel's confidence and visualize it during incremental mapping on the \textit{Hydrant} scene. The surfels in central part of the scene gradually gain higher confidence as observations accumulate, while peripheral surfels remain low.
}
  \label{fig:exp_confidence_update}
\end{figure}

\noindent\textbf{Sparse-to-Dense Tracking.}
We evaluate the impact of the sparse-to-dense strategy on tracking accuracy on the TUM-RGBD dataset. 
The results are shown in Tab.~\ref{tab:tracking_ablation}. 
\note{(Author: tracking ablation)}
\revise{The proposed sparse-to-dense tracking strategy benefits from the high-quality initialization provided by sparse tracking, allowing dense tracking to require minimal optimization. As a result, the poor performance of the \textit{w/o sparse} variant in complex scenes (\textit{e.g.}, \texttt{fr1/room}, \texttt{fr3/office}) is understandable due to the lack of proper initialization.}

\noindent\textbf{Depth-Aware Scale Initialization.}
We compare different surfel scale initialization strategies on the Replica \texttt{off0/off1} in terms of appearance reconstruction quality. 
For the fixed scale initialization strategy, we initialize the size of each surfel ranging from 0.002 to 0.010 with an interval of 0.002 (uint: \textit{cm}). 
For the proposed depth-aware scale initialization strategy, we adjust the scaling factor by tuning $\alpha_s$ ranging from 1.0 to 3.0 with an interval of 0.5. 
As shown in Fig.~\ref{fig:exp_surfel_scale_ablation}, we report the number of surfels and the corresponding PSNR of the final rendered results. 
The proposed initialization strategy significantly improves rendering quality with the same number of surfels.  
To balance efficiency and accuracy, We set \(\alpha_s = 2.0\).

\noindent\textbf{Surfel Fusion with Information Filter.}
We conducted an ablation study on the surfel fusion strategy. As shown in Fig.~\ref{fig:exp_surfel_fusion_ablation}, we performed qualitative analysis on the TUM-RGBD dataset, the proposed fusion strategy integrates multi-view depth observations, effectively mitigating sensor noise and achieving smoother and more consistent scene reconstruction, such as the smooth surfaces of the cup and mouse. For distant objects (\textit{e.g.}, the teddy bear) where depth measurements are more noise-prone, the proposed fusion method demonstrates particularly notable improvements.
\note{(Author: quantitative results of surfels fusion strategy)}
\revise{Then we evaluated its effectiveness in reconstruction tasks on ScanNet++ using the same configuration as in Tab.~\ref{tab:reconstruction}. The results shown in Tab.~\ref{tab:recon_ablation}, demonstrate that the proposed fusion strategy significantly improves the average geometric reconstruction metrics on ScanNet++ scenes.}
\note{(Author: add detail of Fig.~\ref{fig:exp_confidence_update})}
\revise{In addition, we qualitatively visualized the geometric confidence of the \textit{Hydrant} scene from the Azure dataset in Fig.~\ref{fig:exp_confidence_update}. Here, we use the trace of the information matrix $\text{tr}(\Lambda)$, as a proxy for overall confidence, since larger values correspond to lower uncertainty across dimensions. As expected, regions that are continuously observed exhibit higher geometric confidence.}
{\scriptsize  
    \begin{table}
        \caption{Ablation study of geometric regularization.}
        \centering
        \tabcolsep=0.2cm
       \renewcommand\arraystretch{1.05}
        \resizebox{\columnwidth}{!}{
        \begin{tabular}{l c c c c c c}
        \toprule[1pt] 
            Method & ATE$\downarrow$ & PSNR$\uparrow$  & SSIM$\uparrow$   & LPIPS$\downarrow$ & Acc.$\downarrow$  & Comp.$\downarrow$  \\ \hline
            \textit{w/o. reg} 
            & 0.166 & 39.46 & 0.988 & 0.061 & 0.513 & 1.389\\
            \textit{w. reg}
            & \textbf{0.158} & \textbf{39.75} & \textbf{0.991} & \textbf{0.058} & \textbf{0.509} & \textbf{1.384} \\
        \bottomrule[1pt]
        \end{tabular}
        }
        \label{tab:geo_reg_ablation}
    \end{table} 
}

\noindent\textbf{Geometric Regularization.}
We conducted an ablation study on the geometric regularization on Replica \texttt{office0} and report metrics such as trajectory accuracy, rendering quality, and reconstruction quality. As shown in Tab.~\ref{tab:geo_reg_ablation}, it demonstrates that using the geometric regularization term significantly improves camera tracking accuracy as well as the appearance and geometry reconstruction of the scene.

\section{Limitations}
\label{sec:limitations}
\begin{figure}[!h]
  \centering 
  \includegraphics[clip,width=0.98\linewidth, trim={0.0cm 7.8cm 8cm 0cm}]{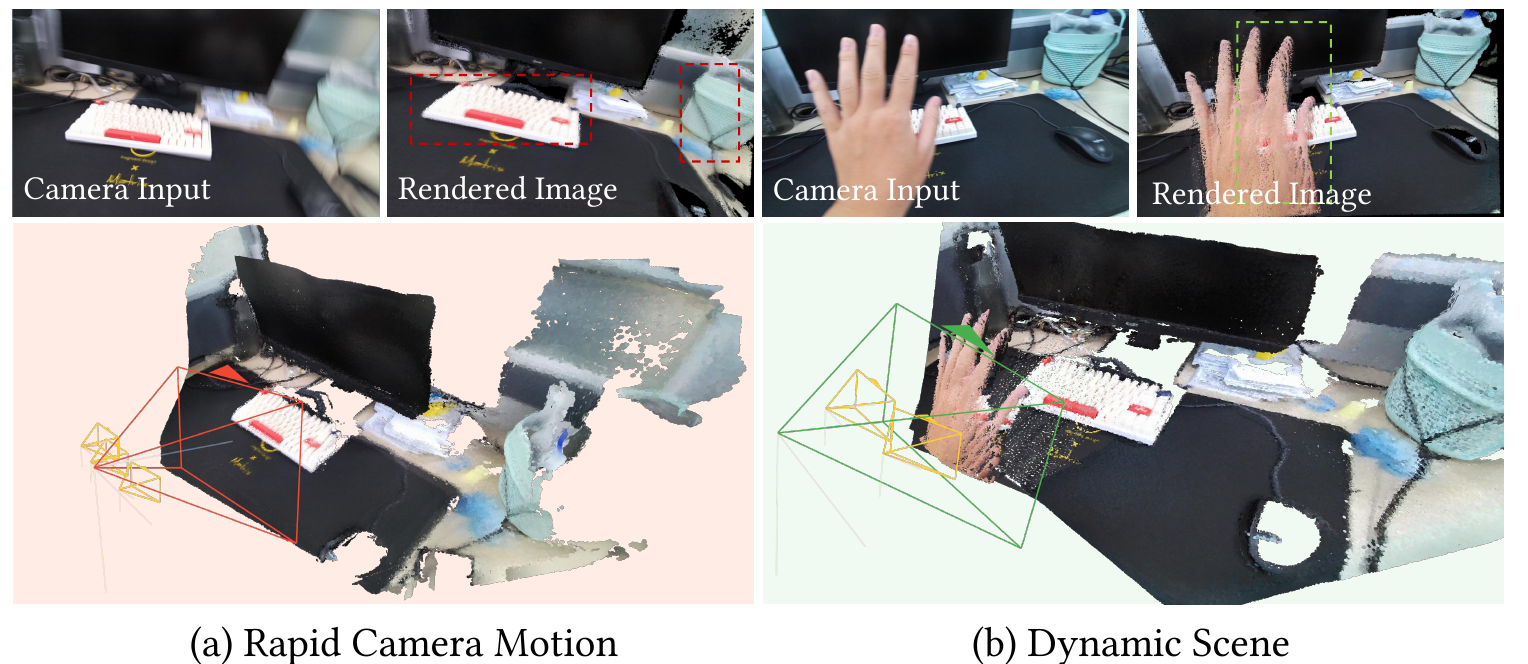}
  \caption{
  \revise{
    Failure case of EGG-Fusion. We present the reconstruction scene and rendered images obtained with EGG-Fusion under (a) rapid camera motion and (b) the presence of dynamic objects in the scene.
  }
  }
  \label{fig:failure_case}
\end{figure}
Although EGG-Fusion demonstrates efficient and stable real-time performance, it still has some limitations. First, real-time mapping using RGB sensors has broader practical applications, 
but the absence of depth measurements imposes higher demands on the robust estimation of surfel geometry positions and normal attributes. 
Second, artifacts caused by certain extreme motion patterns (\textit{e.g.}, wide-baseline localization due to rapid motion or rolling shutter effects) may even lead to tracking failure. 
Finally, our method assumes a static environment, and the presence of dynamic objects or people during scanning may affect reconstruction quality.

\note{(Author: failure case)}
\revise{
As shown in Fig.~\ref{fig:failure_case}, we present two failure cases. Under rapid camera motion, tracking is highly prone to loss, resulting in significant discrepancies between the rendered image and the current frame. In addition, motion blur in such cases leads to the failure of reconstructing details, making it a critical problem to address the degradation of system performance under extreme motion conditions~\cite{seiskari2024gaussian, peng2024bags, lu2025bard}. In dynamic scenes, hand movements introduce ghosting artifacts, which severely affect the rendered image quality. Therefore, an important future direction is to develop robust tracking and reconstruction capabilities for dynamic environments~\cite{dai20254d, lei2025mosca, wu20244d, huang2025rtdepth, zhai2025panogs}.
}

\section{Conclusion}
\label{sec:conclusion}
We propose EGG-Fusion, a robust real-time RGB-D SLAM system that utilizes Gaussian surfels as the map representation, achieving accurate camera tracking and high-quality Gaussian map reconstruction at 24 FPS. This system excels in delivering high geometric precision, superior map rendering quality, and a more compact map representation. Beyond surpassing existing methods on public benchmarks, our system demonstrates exceptional performance in real-world scenarios with live data. 
By enabling efficient object- and scene-level map reconstruction, the proposed real-time 3D reconstruction system offering significant potential for applications in robotics, augmented reality, and autonomous driving.

\begin{acks}
\label{sec:acknowledgment}
This work was partially supported by NSF of China (No. 62425209)
\end{acks}

\bibliographystyle{ACM-Reference-Format}
\bibliography{reference}

\renewcommand{\thesection}{\Alph{section}}
\renewcommand{\thesubsection}{A.\arabic{subsection}}
\section*{Supplementary Materials}


\subsection{Implementation Details}

Our system mainly consists of two modules: the tracking module and the dense mapping module. The tracking module is responsible for preprocessing the input RGB-D frame and performing tracking optimization, including pose estimation based on sparse-correspondences and dense alignment. The dense mapping module handles the initialization of Gaussian surfels, surfel fusion based on information filtering, and end-to-end optimization of geometry and appearance through rasterization.

To achieve overall system efficiency, we employ different implementation strategies tailored to each module. In the data preprocessing stage, we utilize CUDA to efficiently process the input RGB-D data stream, including filtering and the computation of vertex and normal maps. For pose initialization based on sparse correspondences, we adopt the frontend module from ORB-SLAM2~\cite{mur2017orb}, which leverages ORB~\cite{rublee2011orb} features to perform both 2D-2D and 2D-3D matching. This component is implemented in C++ and invoked from the main program via a Python interface. After pose initialization, dense alignment is further applied for pose refinement. This process employs PyTorch-based tensor computations to leverage GPU acceleration for per-pixel matching, local linearization, and reduction. For the differentiable optimization of Gaussian surfels, we build upon the CUDA implementations of ~\cite{daiHighqualitySurfaceReconstruction2024, kerbl3DGaussianSplatting2023}, within which we integrate the functionality of surfel fusion based on information filtering. The main structure of the program is implemented in Python to orchestrate and connect the different modules.

All experiments were conducted on a machine equipped with an RTX 4090 GPU with 24GB of memory and an Intel i9-14900KF CPU with 32 threads.

\subsection{Camera Pose Estimation}
In the dense alignment stage, we employ PyTorch-based tensor computation to leverage GPU acceleration for per-pixel matching, local linearization, and reduction. To solve Eq.(15), we adopt a coarse-to-fine pyramid strategy. Specifically, both the rendered global model surface and the current frame image are downsampled into a multi-scale image pyramid with $L_\text{pyr}$ levels, and each level is optimized for $N_\text{pyr}$ iterations. Starting from the coarsest level, we perform dense alignment using a least-squares method and progressively refine the solution to the original image resolution.

We compute the Jacobian matrix $\textbf{J}$ for both the ICP and photometric residuals, and at each iteration, the update step is computed as:
\begin{align} 
  \delta\boldsymbol{\xi}^{(n)} = -(\textbf{J}^{\top}\textbf{J} + \lambda \textbf{I})^{-1}\textbf{J}^{\top}r(\boldsymbol{\xi}^{n}).
\end{align}
The current estimate is then updated by:
\begin{align} 
  \boldsymbol{\xi}^{n+1} \longleftarrow   \boldsymbol{\xi}^{n} \circ  \delta\boldsymbol{\xi}^{(n)}.
\end{align}
The optimization terminates once the total number of iterations reaches $N_\text{pyr} \cdot L_\text{pyr}$. In our default setting, we use $N_\text{pyr} = 2$ and $L_\text{pyr} = 3$.

\subsection{KeyFrame Selection}
During the tracking process, we determine keyframes to serve as target images for both local and global optimization of the Gaussian surfels map. They also act as optimization targets in the sliding window optimization. The first input frame is set as a keyframe, and subsequent frames are determined as keyframes based on whether the translation $t$ or rotation angle $\theta$ relative to the previous keyframe exceeds a predefined threshold $t_k$ and $\theta_k$. The default setting is $t_k = 0.3m$ and $\theta_k = 20^{\circ}$.

\subsection{Surfels Selection for Fusion}
To determine the set of visible surfels within the current camera frustum, we define \textit{visibility} in a geometric sense, without considering occlusion. Two criteria are used:
1) The projection of the surfel $S_i$ onto the image plane falls within the valid image region.
2) The normal of the surfel is oriented towards the camera.
Formally, the visible surfel set is defined as:
\begin{align} 
  \mathcal{S}^{\text{vis}} = \left \{ S_i \in \mathcal{S} \;\middle|\;  \Pi(\mathbf{p}'_i) \in \Omega \wedge  \textbf{n}_i \cdot \textbf{R}_{t}^z < 0\right \} ,
\end{align}
Here, 
\(\mathbf{p}'_i\) is the center of surfel $S_i$ in the current frame’s coordinate system.
$\Omega = \{ (u, v) \in \mathbb{R}^2 \mid 0 \leq u < W,; 0 \leq v < H \}$
denotes the image coordinate domain, and $\textbf{R}_t^z$ refers to the $z$-axis of the rotation component of the current camera pose.For each $S_i \in \mathcal{S}^{\text{vis}}$, we define the set of surface surfels from the current viewpoint using the following criterion:
\begin{align} 
\mathcal{S}^{\text{surf}} = \left\{ S_i \in \mathcal{S}^{\text{vis}} \;\middle|\; 
\left| 
\left[ \mathbf{p}'_i \right]_z - \bar{D}_t(\mathbf{u}_i)
\right| < \delta_s \right\},
\end{align} 
where $\bar{D}_t$ is the rendered depth from the current viewpoint, and $\delta_s$ denotes the surface thickness threshold, $\textbf{p}_i^{'}$ and $\textbf{u}_i$ have been define in Sec. 3.2.2. Then we check the $\bar{D}_t$ with depth value under current view to identify whether it is re-measured.

\begin{figure}[!h]
  \centering 
  \includegraphics[clip,width=0.98\linewidth, trim={0.0cm 7.cm 15cm 0cm}]{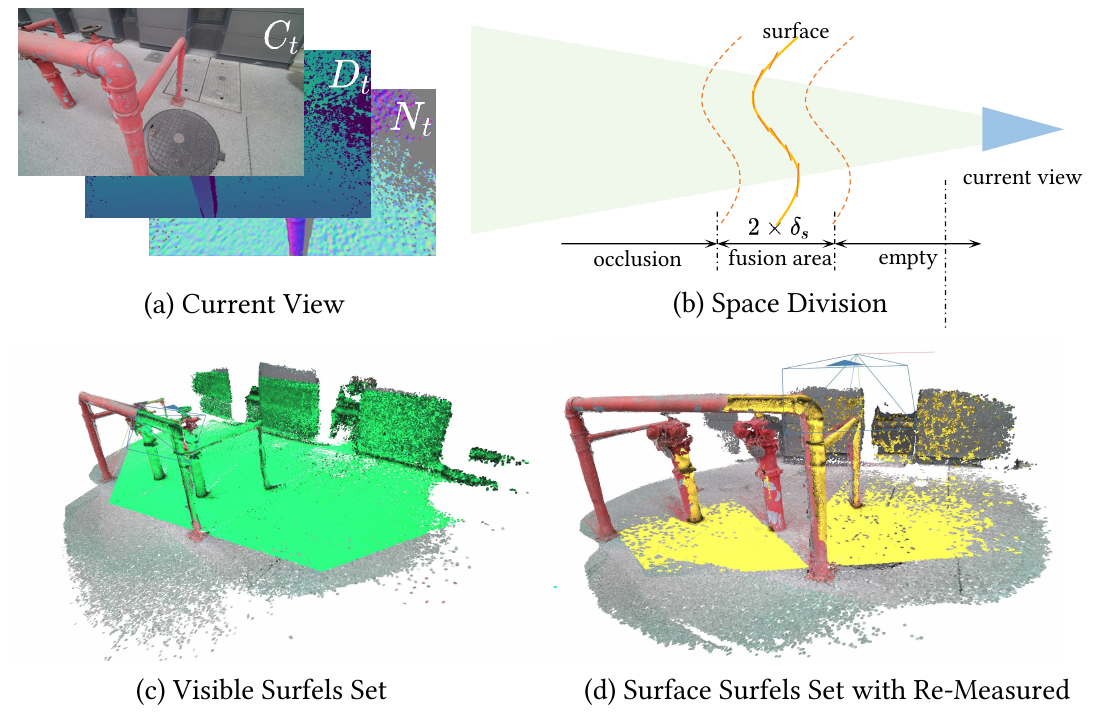}
  \caption{Surfels Selection for Fusion}
  \label{fig:surface_vis}
\end{figure}

\subsection{Detail of Rotation Matrix Update}
Regarding the details of Eq.(8), we omit the explicit form of the rotation matrix conversion from a rotation vector, denoted as $\Delta \mathbf{R}(\mathbf{n},\theta)$. In our implementation, we adopt the cross-product form of the Rodrigues' rotation formula:
\begin{align} 
  \Delta \mathbf{R}(\mathbf{n},\theta) = \cos\theta \cdot \mathbf{I} + (1 - \cos\theta) \cdot \mathbf{n} \mathbf{n}^\top + \sin\theta \cdot [\mathbf{n}]_\times 
\end{align}
Here, $[\mathbf{n}]_\times$ denotes the skew-symmetric matrix of a vector $\mathbf{n}$:
\begin{align} 
[\mathbf{n}]_\times =
\begin{bmatrix}
0 & -n_z & n_y \\
n_z & 0 & -n_x \\
-n_y & n_x & 0
\end{bmatrix}
\end{align}

\subsection{Running Time on Replica}

\begin{figure}[!h]
  \centering 
  \includegraphics[clip,width=0.98\linewidth, trim={0.0cm 7cm 12cm 0cm}]{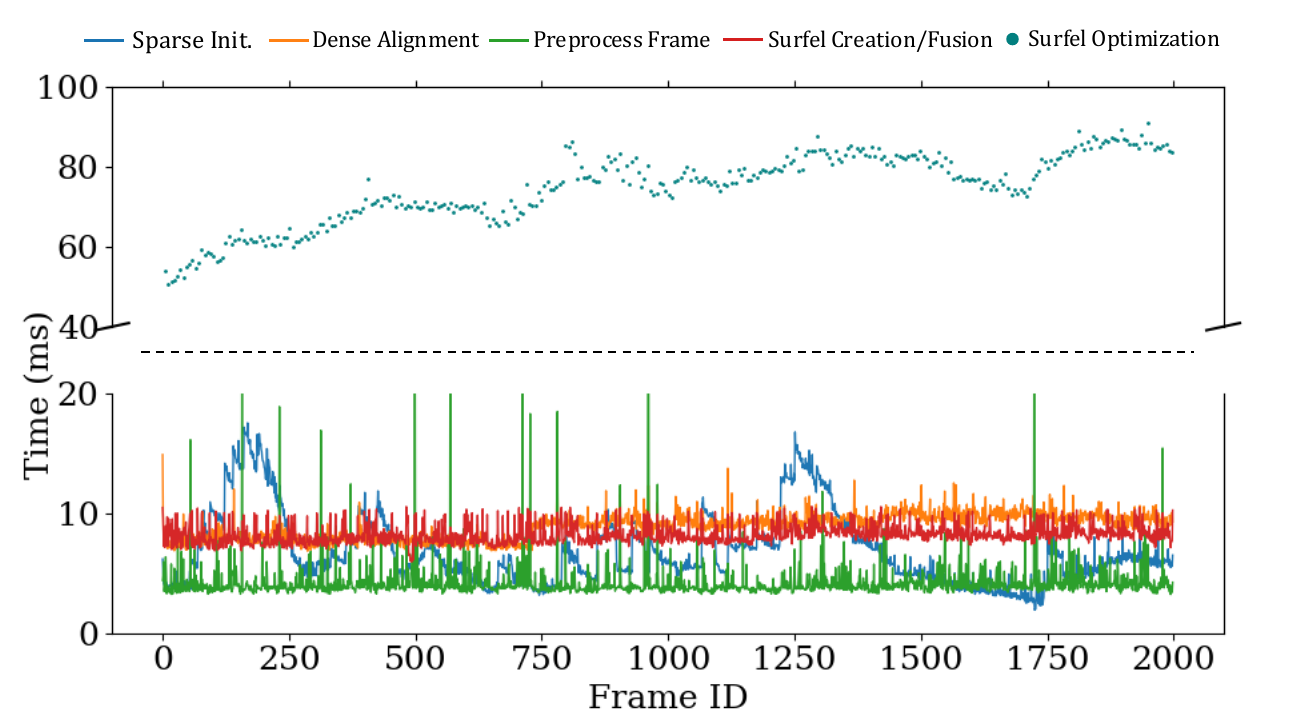}
  \caption{Runtime of our system. We plotted the time consumption of the main modules of the system  on Replica \texttt{office0}.}
  \label{fig:exp_system_performance}
\end{figure}

We report the time consumption for each frame in the major modules on the \texttt{office0}, as shown in the Fig.~\ref{fig:exp_system_performance}. We provide statistics for the camera tracking, which includes sparse-correspondence-based pose initialization and dense alignment. Each frame undergoes preprocessing to prepare data for the mapping stage. In the mapping stage, each frame contributes to adding new surfels to the global map and fusing them with existing surfels. Finally, we optimize the Gaussian surfels using frame batches at a certain frequency. This part is comparatively more time-consuming than the other modules but the is still significantly less than that of other methods.

\subsection{Meshing with Voxel Masking}

\begin{figure}[!h]
  \centering 
  \includegraphics[clip,width=0.98\linewidth, trim={0.0cm 7.5cm 11cm 0cm}]{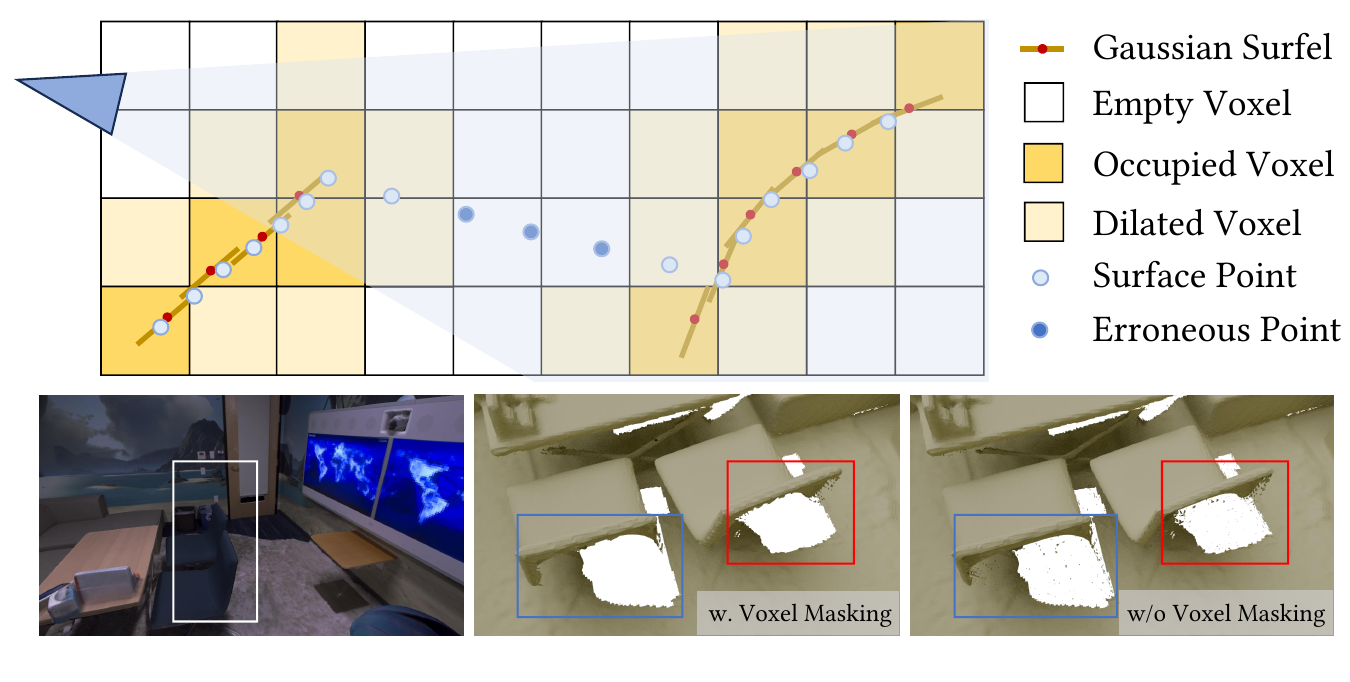}
  \caption{Meshing the Gaussian Surfel based Scene Map with Voxel Masking}
  \label{fig:voxel_masking}
\end{figure}

\begin{figure*}[!h]
  \centering 

\begin{minipage}{0.02\linewidth} 
    \centering
    \small
    \parbox[c][42pt][c]{\textwidth}{\centering \rotatebox{90}{\texttt{Scene1}}} \\
    \parbox[c][42pt][c]{\textwidth}{\centering \rotatebox{90}{\texttt{Scene2}}} \\
    \parbox[c][42pt][c]{\textwidth}{\centering \rotatebox{90}{\texttt{Scene3}}} \\
    \parbox[c][42pt][c]{\textwidth}{\centering \rotatebox{90}{\texttt{Scene4}}}
\end{minipage}%
\raisebox{-0.0\height}{%
  \begin{minipage}[c]{0.98\linewidth}
    \includegraphics[clip,width=1\linewidth, trim={1.1cm 1.05cm 0cm 0cm}]{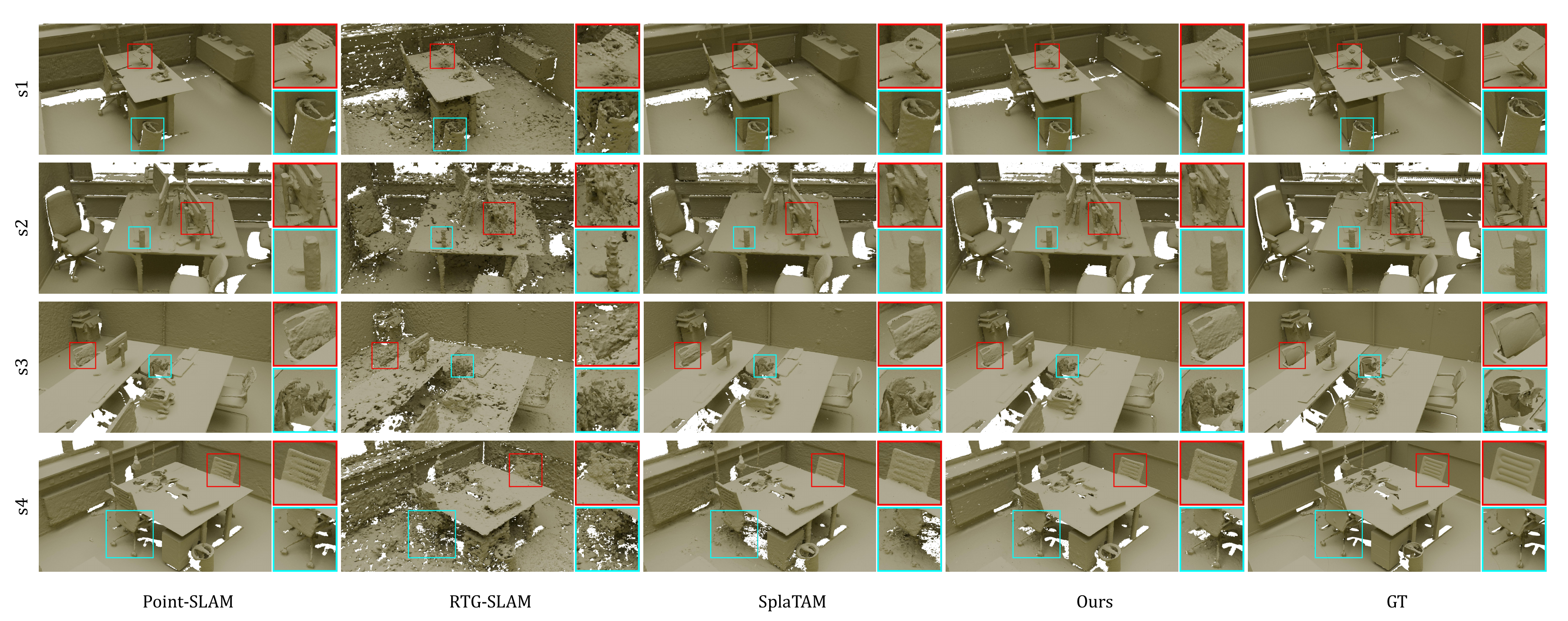}
  \end{minipage}
}

\begin{minipage}{0.05\linewidth}
 \centering
 \fontsize{7pt}{8pt}\selectfont
 \textcolor{white}{x}
 \end{minipage}
 \hfill
 \begin{minipage}{0.18\linewidth}
 \centering
 \fontsize{7pt}{8pt}\selectfont
 Point-SLAM~\shortcite{point_slam}
 \end{minipage}
 \hfill
 \begin{minipage}{0.16\linewidth}
 \centering
 \fontsize{7pt}{8pt}\selectfont
 RTG-SLAM~\shortcite{rtg_slam}
 \end{minipage}
 \hfill
  \begin{minipage}{0.16\linewidth}
 \centering
 \small
 SplaTAM~\shortcite{splatam}
 \end{minipage}
 \hfill
 \begin{minipage}{0.16\linewidth}
 \centering
 \fontsize{7pt}{8pt}\selectfont
 EGG-Fusion(Ours)
 \end{minipage}
  \hfill
  \begin{minipage}{0.16\linewidth}
 \centering
 \fontsize{7pt}{8pt}\selectfont
 GT
 \end{minipage}
\hfill
\begin{minipage}{0.05\linewidth}
\centering
\fontsize{7pt}{8pt}\selectfont
\textcolor{white}{x}
\end{minipage}

  \caption{Reconstruction results on ScanNet++ dataset.}
  \label{fig:scannetpp_mesh_full}
\end{figure*}

\begin{table*}[t]
    \caption{Comparison of train view synthesis on Replica. $^{*}$ indicates that the result is taken from~\cite{zhuNICERSLAMNeuralImplicit2023}}
    \centering
    \scriptsize
    \tabcolsep=0.4cm
    \renewcommand\arraystretch{1.2}
    \begin{tabular}{l l c c c c c c c c c}
    \toprule[1pt]
        Method     & Metric               & \texttt{Rm 0}   & \texttt{Rm 1}  & \texttt{Rm 2}  & \texttt{Off 0} & \texttt{Off 1} & \texttt{Off 2} & \texttt{Off 3} & \texttt{Off 4} & \texttt{Avg.} \\ \hline
        \multirow{3}{*}{\makecell{NICE-SLAM$^{*}$~\shortcite{nice_slam}\\}} 
                    &PSNR$\uparrow$      & 22.12  & 22.47 & 24.52 & 29.07 & 30.34 & 19.66 & 22.23 & 24.94 & 24.42 \\
                    &SSIM$\uparrow$      & 0.689  & 0.757 & 0.814 & 0.874 & 0.886 & 0.797 & 0.801 & 0.856 & 0.809 \\
                    &LPIPS$\downarrow$   & 0.330  & 0.271 & 0.208 & 0.229 & 0.181 & 0.235 &  0.209 & 0.198 & 0.233 \\ \hline
        \multirow{3}{*}{\makecell{Vox-Fusion~\shortcite{vox_fusion}\\}} 
                    &PSNR$\uparrow$     & 26.16 & 28.16 & 28.03 & 32.49 & 32.45 & 26.86 & 27.27 & 29.61 & 28.88 \\
                    &SSIM$\uparrow$     & 0.898 & 0.904 & 0.918 & 0.942 & 0.952 & 0.934 & 0.941 & 0.946 & 0.929 \\
                    &LPIPS$\downarrow$  & 0.293 & 0.271 & 0.232 & 0.216 & 0.207 & 0.227 & 0.187 & 0.199 & 0.229 \\ \hline
        \multirow{3}{*}{\makecell{Point-SLAM~\shortcite{point_slam}\\}} 
                    &PSNR$\uparrow$      & \cellcolor{red!40}32.40 & \cellcolor{orange!40}34.08 & \cellcolor{red!40}35.50 & \cellcolor{yellow!40}38.26 & \cellcolor{yellow!40}39.16 & \cellcolor{red!40}33.99 & \cellcolor{red!40}33.48 & \cellcolor{yellow!40}33.49 & \cellcolor{orange!40}35.17 \\
                    &SSIM$\uparrow$      & \cellcolor{red!40}0.974 & \cellcolor{orange!40}0.977 & \cellcolor{orange!40}0.982 & \cellcolor{yellow!40}0.983 & \cellcolor{yellow!40}0.986 & 0.960 & \cellcolor{yellow!40}0.960 & \cellcolor{yellow!40}0.979 & \cellcolor{yellow!40}0.975 \\
                    &LPIPS$\downarrow$   & \cellcolor{orange!40}0.113 & 0.116 & \cellcolor{yellow!40}0.111 & 0.100 & 0.118 & 0.156 & \cellcolor{yellow!40}0.132 & \cellcolor{yellow!40}0.142 & 0.124 \\ \hline
        \multirow{3}{*}{\makecell{SplaTAM~\shortcite{splatam}\\}} 
                    &PSNR$\uparrow$      & \cellcolor{orange!40}32.09 & \cellcolor{yellow!40}33.62 & \cellcolor{orange!40}35.00 & 38.16 & 39.04 & 31.88 & 30.14 & 31.69 & 33.95 \\
                    &SSIM$\uparrow$      & \cellcolor{orange!40}0.972 & 0.970 & \cellcolor{orange!40}0.982 & 0.982 & 0.982 & \cellcolor{yellow!40}0.965 & 0.950 & 0.947 & 0.969 \\
                    &LPIPS$\downarrow$   & 0.077 & \cellcolor{orange!40}0.097 & \cellcolor{red!40}0.074 & \cellcolor{yellow!40}0.086 & \cellcolor{orange!40}0.092 & \cellcolor{red!40}0.099 & \cellcolor{orange!40}0.118 & 0.155 & \cellcolor{orange!40}0.100 \\ \hline
        \multirow{3}{*}{\makecell{RTG-SLAM~\shortcite{rtg_slam}\\}} 
                    &PSNR$\uparrow$      & 30.91 & 33.41 & 34.49 & \cellcolor{orange!40}39.02 & \cellcolor{orange!40}39.24 & \cellcolor{yellow!40}32.78 & \cellcolor{yellow!40}32.73 & \cellcolor{orange!40}35.56 & \cellcolor{yellow!40}34.76 \\
                    &SSIM$\uparrow$      & 0.962 & \cellcolor{yellow!40}0.976 & 0.981 & \cellcolor{orange!40}0.989 & \cellcolor{orange!40}0.989 & \cellcolor{orange!40}0.980 & \cellcolor{red!40}0.981 & \cellcolor{orange!40}0.984 & \cellcolor{orange!40}0.980 \\
                    &LPIPS$\downarrow$   & \cellcolor{red!40}0.147 & \cellcolor{yellow!40}0.116 & 0.120 & \cellcolor{orange!40}0.082 & \cellcolor{yellow!40}0.099 & \cellcolor{yellow!40}0.144 & 0.138 & \cellcolor{orange!40}0.123 & \cellcolor{yellow!40}0.121 \\ \hline
        \multirow{3}{*}{\makecell{Ours}}
                    &PSNR$\uparrow$      & \cellcolor{yellow!40}31.21 & \cellcolor{red!40}34.14 & \cellcolor{yellow!40}34.94 & \cellcolor{red!40}39.75 & \cellcolor{red!40}39.69 & \cellcolor{orange!40}32.98 & \cellcolor{orange!40}33.14 & \cellcolor{red!40}35.95 & \cellcolor{red!40}35.23 \\
                    &SSIM$\uparrow$      & \cellcolor{yellow!40}0.966 & \cellcolor{red!40}0.979 & \cellcolor{red!40}0.983 & \cellcolor{red!40}0.990 & \cellcolor{red!40}0.991 & \cellcolor{red!40}0.983 & \cellcolor{orange!40}0.980 & \cellcolor{red!40}0.988 & \cellcolor{red!40}0.983 \\
                    &LPIPS$\downarrow$   & \cellcolor{yellow!40}0.131 & \cellcolor{red!40}0.089 & \cellcolor{orange!40}0.098 & \cellcolor{red!40}0.058 & \cellcolor{red!40}0.065 & \cellcolor{orange!40}0.111 & \cellcolor{red!40}0.106 & \cellcolor{red!40}0.108 & \cellcolor{red!40}0.096 \\
    \bottomrule[1pt]
    \end{tabular}
    \label{tab:replica_rendering}
\end{table*}

As emphasized in~\cite{daiHighqualitySurfaceReconstruction2024}, the alpha decay property based on surfel ellipsoid centers leads to erroneous depth estimates in regions with depth discontinuities. Consequently, when meshing the Gaussian surfel based map, the edges often contain noisy or scattered triangles. ~\cite{daiHighqualitySurfaceReconstruction2024} addresses this issue using a volumetric cutting strategy to prevent such artifacts during screened Poisson surface reconstruction.
Similarly, when we follow the TSDF-based meshing approaches in~\cite{nice_slam, vox_fusion}, we encounter the same problem. Therefore, we adopt a voxel masking strategy suitable for voxel-based surface extraction.

As illustrated in Fig.~\ref{fig:voxel_masking}, we construct an occupancy grid of the scene based on the surfel centers, and then determine whether points from the rendered depth map should contribute to TSDF integration by checking if they fall within the occupied voxels. This is based on the fact that the scene surfaces generally lie within the spatial vicinity of the surfels.

To mitigate the quantization errors introduced by voxelization where points may be near surfels but still fall into empty voxels, we further apply voxel dilation. This ensures more accurate selection of surface points for TSDF-based meshing.

\subsection{Rendering Results on Replica}

The quality of view synthesis under training views across 8 scenes in the Replica dataset is shown in Tab.~\ref{tab:replica_rendering}. Among NeRF-based and GS-based methods, our method achieved the best average rendering quality and the best rendering quality in most scenes. We believe this is due to the Gaussian surfel tightly fitting the scene surface with the geometric regularization while end-to-end optimization.

\subsection{Detail of Reconstruction Results on ScanNet++}

\begin{table*}[t]
\centering
\scriptsize
\tabcolsep=0.37cm
\renewcommand{\arraystretch}{1.2} 

\caption{Comparison of rendering capabilities on ScanNet++, including both training views and test views (views not seen during training).}

\begin{tabular}{lcccccc|ccccc}
\toprule

\multirow{2}{*}{\textbf{Methods}} & \multirow{2}{*}{\textbf{Metrics}} & \multicolumn{5}{c}{\textbf{Novel View}} & \multicolumn{5}{c}{\textbf{Training View}} \\

 &  & \textbf{Avg.} & \texttt{S1} & \texttt{S2} & \texttt{S3} & \texttt{S4} & \textbf{Avg.} & \texttt{S1} & \texttt{S2} & \texttt{S3} & \texttt{S4} \\

\midrule

\multirow{3}{*}{\makecell{Point-SLAM~\shortcite{point_slam}\\}}
& PSNR $\uparrow$ 
& 17.68 & 15.00 & 21.63 & 15.87 & 18.21 
& 24.35 & 24.71 & 23.13 & 24.77 & 24.79 \\
& SSIM $\uparrow$ 
& 0.623 & 0.611 & 0.702 & 0.560 & 0.618 
& 0.800 & 0.805 & 0.783 & 0.813 & 0.800 \\
& LPIPS $\downarrow$ 
& 0.548 & 0.558 & 0.480 & 0.614 & 0.539 
& 0.373 & 0.367 & 0.383 & 0.375 & 0.366 \\

\midrule

\multirow{3}{*}{\makecell{SplaTAM~\shortcite{splatam}\\}}
& PSNR $\uparrow$ 
& 24.75 & 24.08 & \cellcolor{orange!40}26.41 & 25.19 & \cellcolor{orange!40}23.33 
& 27.30 & 27.82 & \cellcolor{orange!40}25.42 & 28.22 & 27.75 \\
& SSIM $\uparrow$ 
& \cellcolor{orange!40}0.900 & \cellcolor{orange!40}0.886 & \cellcolor{red!40}0.930 & \cellcolor{orange!40}0.888 & \cellcolor{orange!40}0.897 
& \cellcolor{orange!40}0.940 & \cellcolor{orange!40}0.946 & \cellcolor{red!40}0.924 & \cellcolor{orange!40}0.943 & \cellcolor{red!40}0.947 \\
& LPIPS $\downarrow$ 
& \cellcolor{red!40}0.208 & \cellcolor{orange!40}0.211 & \cellcolor{red!40}0.175 & \cellcolor{orange!40}0.253 & \cellcolor{orange!40}0.195 
& \cellcolor{red!40}0.130 & \cellcolor{red!40}0.119 & \cellcolor{red!40}0.158 & \cellcolor{red!40}0.130 & \cellcolor{red!40}0.112 \\

\midrule

\multirow{3}{*}{\makecell{RTG-SLAM~\shortcite{rtg_slam}\\}}
& PSNR $\uparrow$ 
& \cellcolor{orange!40}24.77 & \cellcolor{orange!40}24.27 & 25.44 & \cellcolor{orange!40}26.09 & 23.28 
& \cellcolor{orange!40}27.54 & \cellcolor{orange!40}28.22 & 24.69 & \cellcolor{orange!40}29.29 & \cellcolor{orange!40}27.96 \\
& SSIM $\uparrow$ 
& 0.882 & 0.876 & 0.886 & 0.883 & 0.882 
& 0.925 & 0.936 & 0.889 & 0.937 & 0.936\\
& LPIPS $\downarrow$ 
& 0.255 & 0.249 & 0.261 & 0.285 & 0.225
& 0.184 & 0.165 & 0.238 & 0.176 & 0.155 \\

\midrule

\multirow{3}{*}{\textbf{Ours}}
& PSNR $\uparrow$ 
& \cellcolor{red!40}25.70 & \cellcolor{red!40}25.50 & \cellcolor{red!40}26.55 & \cellcolor{red!40}26.72 & \cellcolor{red!40}23.96
& \cellcolor{red!40}29.06 & \cellcolor{red!40}29.97 & \cellcolor{red!40}26.08 & \cellcolor{red!40}30.59 & \cellcolor{red!40}29.45 \\
& SSIM $\uparrow$ 
& \cellcolor{red!40}0.907 & \cellcolor{red!40}0.906 & \cellcolor{orange!40}0.901 & \cellcolor{red!40}0.900 & \cellcolor{red!40}0.903
& \cellcolor{red!40}0.944 & \cellcolor{red!40}0.953 & \cellcolor{orange!40}0.922 & \cellcolor{red!40}0.950 & \cellcolor{orange!40}0.943 \\
& LPIPS $\downarrow$ 
& \cellcolor{orange!40}0.212 & \cellcolor{red!40}0.196 & \cellcolor{orange!40}0.231 & \cellcolor{red!40}0.250 & \cellcolor{red!40}0.190
& \cellcolor{orange!40}0.141 & \cellcolor{orange!40}0.121 & \cellcolor{orange!40}0.178 & \cellcolor{orange!40}0.145 & \cellcolor{orange!40}0.116 \\

\bottomrule
\end{tabular}
\label{tab:scannetpp_rendering}
\end{table*}

Due to space constraints in the article, we only presented the average rendering metrics on ScanNet++~\cite{scannet++}. 
As shown in Tab.~\ref{tab:scannetpp_rendering}, we provide the complete results for a more comprehensive analysis.

\subsection{More Results}
\begin{figure*}[!h]
  \centering 
\begin{minipage}[c]{0.02\linewidth} 
    \centering
    \small
    \parbox[c][52pt][c]{\textwidth}{\centering \rotatebox{90}{\texttt{Office0}}} \\
    \parbox[c][52pt][c]{\textwidth}{\centering \rotatebox{90}{\texttt{Office1}}} \\
    \parbox[c][52pt][c]{\textwidth}{\centering \rotatebox{90}{\texttt{Office2}}} \\
    \parbox[c][52pt][c]{\textwidth}{\centering \rotatebox{90}{\texttt{Office3}}} \\
    \parbox[c][52pt][c]{\textwidth}{\centering \rotatebox{90}{\texttt{Office4}}} \\
    \parbox[c][52pt][c]{\textwidth}{\centering \rotatebox{90}{\texttt{Room1}}} \\
    \parbox[c][52pt][c]{\textwidth}{\centering \rotatebox{90}{\texttt{Room2}}} 
\end{minipage}%
\begin{minipage}[c]{0.98\linewidth}
    \includegraphics[clip,width=1\linewidth, trim={1.3cm 1.35cm 0cm 0cm}]{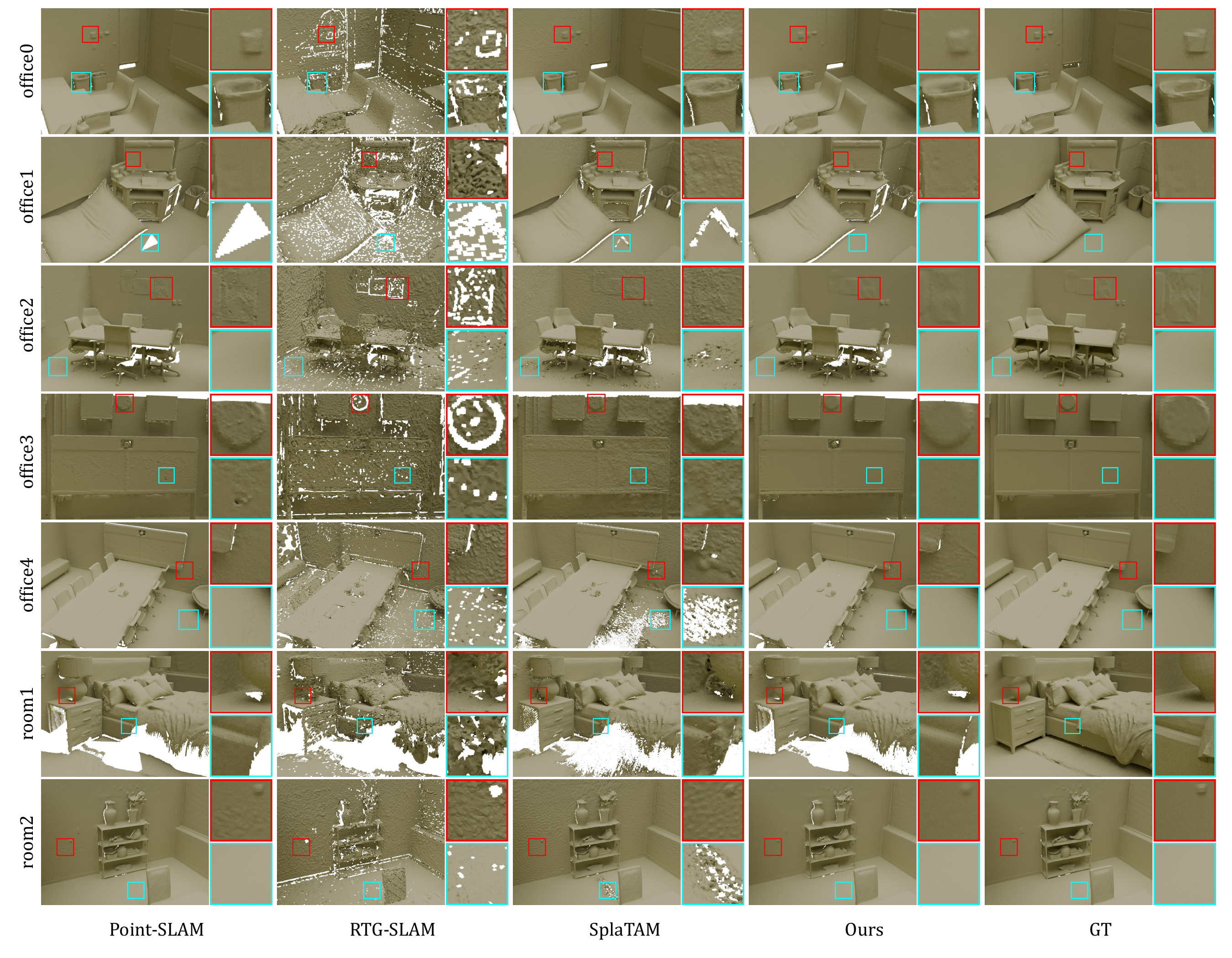}
\end{minipage}

\begin{minipage}{0.02\linewidth}
 \centering
 \fontsize{7pt}{8pt}\selectfont
 \textcolor{white}{x}
 \end{minipage}
 \hfill
 \begin{minipage}{0.18\linewidth}
 \centering
 \fontsize{7pt}{8pt}\selectfont
 Point-SLAM~\shortcite{point_slam}
 \end{minipage}
 \hfill
 \begin{minipage}{0.16\linewidth}
 \centering
 \fontsize{7pt}{8pt}\selectfont
 RTG-SLAM~\shortcite{rtg_slam}
 \end{minipage}
 \hfill
  \begin{minipage}{0.16\linewidth}
 \centering
 \small
 SplaTAM~\shortcite{splatam}
 \end{minipage}
 \hfill
 \begin{minipage}{0.16\linewidth}
 \centering
 \fontsize{7pt}{8pt}\selectfont
 EGG-Fusion(Ours)
 \end{minipage}
  \hfill
  \begin{minipage}{0.16\linewidth}
 \centering
 \fontsize{7pt}{8pt}\selectfont
 GT
 \end{minipage}
\hfill
\begin{minipage}{0.02\linewidth}
\centering
\fontsize{7pt}{8pt}\selectfont
\textcolor{white}{x}
\end{minipage}
  
  \caption{Reconstruction results on Replica dataset.}
  \label{fig:replica_mesh_full}
\end{figure*}

We present the complete mesh reconstruction results on Replica~\cite{straubReplicaDatasetDigital2019} and ScanNet++~\cite{scannet++} to demonstrate the superiority of our method in both reconstruction detail and overall accuracy, as shown as in Fig.~\ref{fig:replica_mesh_full} and Fig.~\ref{fig:scannetpp_mesh_full}.

\end{document}